\documentclass[review]{elsarticle}
\usepackage{lineno}
\usepackage{placeins}
\modulolinenumbers[5]

\usepackage[table]{xcolor}
\usepackage{amssymb}
\usepackage{graphicx,pdfpages}
\usepackage{epsfig}
\usepackage{epstopdf}
\usepackage{times}
\usepackage{bm}
\usepackage{booktabs}
\usepackage{algorithm}
\usepackage{algpseudocode}
\usepackage{url}
\usepackage{amsmath,amsthm}
\usepackage{multirow}
\usepackage{colortbl}
\usepackage{footmisc}
\usepackage{subcaption}
\usepackage{caption}
\usepackage{comment}
\usepackage{color}

\DeclareMathOperator*{\argmax}{arg\,max}
\newcommand{\comments}[1]{}
\emergencystretch=\maxdimen
\newcommand{\sysname}[1]{\textsc{#1}}
\newcommand{\word}[1]{\texttt{#1}}

\begin{document}
\begin{frontmatter}

\title{Latent Tree Models for Hierarchical Topic Detection}
\author[mymainaddress]{Peixian Chen}

\author[mymainaddress]{Nevin L.~Zhang\corref{mycorrespondingauthor}}
\cortext[mycorrespondingauthor]{Corresponding author}
\ead{lzhang@cse.ust.hk}

\author[mysecondaryaddress] {Tengfei Liu} %\email liutf@ust.hk\\

\author[mythirdaddress]{Leonard K.~M. Poon} % \email kmpoon@ied.edu.hk\\

\author[mymainaddress]{Zhourong Chen}

\author[mymainaddress]{Farhan Khawar}

\address[mymainaddress]{Department of Computer Science and Engineering \\
        The Hong Kong University of Science and Technology, Hong Kong}
\address[mysecondaryaddress]{Ant Financial Services Group, Shanghai}

\address[mythirdaddress]{Department of Mathematics and Information Technology\\
        The  Education University of Hong Kong, Hong Kong}

\begin{abstract}
We present a novel method for hierarchical topic detection where topics are obtained by clustering documents in multiple ways. Specifically, we model document collections using a class of graphical models called {\it hierarchical latent tree models (HLTMs)}. The variables at the bottom level of an HLTM are observed binary variables that represent the presence/absence of words in a document. The variables at other levels are binary latent variables, with those at the lowest latent level representing word co-occurrence patterns and those at higher levels representing co-occurrence of patterns at the level below. Each latent variable gives a soft partition of the documents, and document clusters in the partitions are interpreted as topics. Latent variables at high levels of the hierarchy capture long-range word co-occurrence patterns and hence give thematically more general topics, while those at low levels of the hierarchy capture short-range word co-occurrence patterns and give thematically more specific topics. Unlike LDA-based topic models,  HLTMs do not refer to a document generation process and use word variables instead of token variables. They use a tree structure to model the relationships between topics and words, which is conducive to the discovery of meaningful topics and topic hierarchies.
\end{abstract}

\end{frontmatter}

\section{Introduction}
\label{sec.intro}

The objective of {\it hierarchical topic detection (HTD)} is to, given a corpus of documents, obtain a tree of topics with more general topics at high levels of the tree and more specific topics at low levels of the tree. It has a wide range of potential applications. For example, a topic hierarchy for posts at an online forum can provide an overview of the variety of the posts and guide readers quickly to the posts of interest. A topic hierarchy for the reviews and feedbacks on a business/product can help a company gauge customer sentiments and identify areas for improvements. A topic hierarchy for recent papers published at a conference or journal can give readers a global picture of recent trends in the field. A topic hierarchy for all the articles retrieved from PubMed on an area of medical research can help researchers get an overview of past studies in the area. In applications such as those mentioned here, the problem is not about search because the user does not know what to search for. Rather the problem is about summarization of thematic contents and topic-guided browsing.

Several HTD methods have been proposed previously, including nested Chinese  restaurant process (nCRP) \citep{griffiths2004hierarchical, blei10nested}, Pachinko allocation model (PAM) \citep{li2006pachinko,mimno2007mixtures}, and  nested hierarchical Dirichlet process (nHDP)~\citep{paisley2015nested}.  Those methods are extensions of {\em latent Dirichlet allocation (LDA)}~\citep{blei2003latent}. Hence we refer to them collectively as {\em LDA-based methods}.

In this paper, we present a novel HTD method called {\em hierarchical latent tree analysis (HLTA)}. Like the LDA-based methods, HLTA is a probabilistic method and it involves latent variables. However, there are fundamental differences.
The first difference lies in what is being modeled and the semantics of the latent variables. The LDA-based methods model the process by which documents are generated. The latent variables in the models are constructs in the hypothetical generation process, including a list of topics (usually denoted as $\beta$), a topic distribution vector for each document (usually denoted as $\theta_d$), and a topic assignment for each token in each document (usually denoted as $Z_{d, n}$). In contrast,  HLTA models a collection of documents without referring to a document generation process. The latent variables in the model are considered unobserved attributes of the documents. If we compare whether words occur in particular documents to whether students do well in various subjects, then the latent variables correspond to latent traits such as analytical skill, literacy skill and general intelligence.

The second difference lies in the types of observed variables used in the models.
Observed variables in the LDA-based methods are token variables (usually denoted as $W_{d, n}$). Each token variable stands for a location in a document, and its possible values are the words in a vocabulary. Here one cannot talk about conditional independence between words because the probabilities of all words must sum to 1.
In contrast, each observed variable in HLTA stands for a word. It is a binary variable and represents the presence/absence of the word in a document. The output of HLTA is a tree-structured graphical model, where the word variables are at the leaves and the latent variables are at the internal nodes. Two word variables are conditionally independent given any latent variable on the path between them. Words that frequently co-occur in documents tend to be located in the same ``region" of the tree. This fact is conducive to the discovery of meaningful topics and topic hierarchies. A drawback of using binary word variables is that word counts cannot be taken into consideration.

The third difference lies in the definition and characterization of topics.
Topics in the LDA-based methods are probabilistic distributions over a vocabulary. When presented to users, a topic is characterized using a few words with the highest probabilities. In contrast, topics in HLTA are clusters of documents. More specifically, all latent  variables in HTLA are assumed to be binary. Just as the concept ``analytical skill" partitions a student population into soft two clusters, those with high analytic skill in one cluster and those with low analytic skill in another, a latent variable in HLTA partitions a document collection into two soft clusters of documents. The document clusters are interpreted as topics. For presentation to users, a topic is characterized using the words that not only occur with high probabilities in topic but also occur with low probabilities outside the topic. The consideration of occurrence probabilities outside the topic is important because a word that occurs with high probability in the topic might also occur with high probability outside the topic. When that happens, it is not a good choice for the characterization of the topic.

There are other differences that are more technical in nature and  the explanations are hence postponed to Section
\ref{sec.hltm}.

 The rest of the paper is organized as follows. We discuss related work in Section \ref{sec.related} and review the basics of latent tree models in Section \ref{sec.ltm}. In Section  \ref{sec.hltm}, we  introduce hierarchical latent tree models (HLTMs) and explain how they can be used for hierarchical topic detection. The HLTA algorithm for learning HLTMs is described in Sections \ref{sec.modelconstruct} - \ref{sec.big}. In Section
\ref{sec.results}, we present the results HTLA obtains on a real-world dataset and discuss some practical issues.  In Section \ref{sec.comparisonWLDA}, we empirically compare HLTA with the LDA-based methods. Finally, we end the paper in Section~\ref{sec.conclusion} with some concluding remarks and discussions of future work.

\section{Related Work}
\label{sec.related}

Topic detection has been one of the most active research areas in Machine Learning in the past decade. The most commonly used method is latent Dirichlet allocation (LDA)~\cite{blei2003latent}. LDA assumes that documents are generated as follows:
First, a list $\{\beta_1, \ldots, \beta_K\}$ of topics is drawn from a Dirichlet distribution. Then, for each document $d$, a topic distribution  $\theta_d$ is drawn from another Dirichlet distribution. Each word $W_{d, n}$ in the document is generated by
first picking a topic $Z_{d, n}$ according to the topic distribution $\theta_d$, and then selecting a word according to the word distribution $\beta_{Z_{d, n}}$ of the topic.  Given a document collection, the generation process is reverted via statistical inference (sampling or variational inference) to determine the topics and topic compositions of the documents.

LDA has been extended in various ways for additional modeling capabilities.
Topic correlations are considered in~\cite{lafferty2006correlated,li2006pachinko}; topic evolution is modeled in~\cite{blei2006dynamic,wang2006topics,ahmed2007dynamic}; topic structures are built in ~\cite{teh2006hierarchical,li2006pachinko,griffiths2004hierarchical,mimno2007mixtures}; side information is exploited in~\cite{andrzejewski2009,jagarlamudi2012};
supervised topic models are proposed in~\cite{mcauliffe2008supervised,perotte2011hierarchically}; and so on. In the following, we discuss in more details three of the extensions that are more closely related to this paper than others.

 Pachinko allocation model (PAM) \citep{li2006pachinko,mimno2007mixtures} is proposed as a method for modeling correlations among topics. It introduces multiple levels of supertopics on top of the basic topics. Each supertopic is a distribution over the  topics at the next level below. Hence PAM can also be viewed as an HTD method, and the hierarchical structure needs to be predetermined. To pick a topic for a token, it first draws a top-level topic from a multinomial distribution (which in turn is drawn from a Dirichlet distribution), and then draws a topic for the next level below  from the multinomial distribution associated with the top-level topic, and so on. The rest of the generation process is the same as in LDA.

Nested Chinese Restaurant Process (nCRP)~\cite{blei10nested} and nested Hierarchical Dirichlet Process (nHDP)~\cite{paisley2015nested} are proposed as HTD methods.  They assume that there is a true topic tree behind data. A prior distribution is placed over all possible trees using nCRP and nHDP respectively. An assumption is made as to how documents are generated from the true topic tree, which, together with data, gives a likelihood function over all possible trees. In nCRP, the topics in a document are assumed to be from one path down the tree, while in  nHDP, the topics in a document can be from multiple paths, i.e., a subtree within the entire topic tree.
 The true topic tree is estimated by combining the prior and the likelihood in posterior inference. During inference, one in theory deals with a tree with infinitely many levels and each node having infinitely many children. In practice, the tree is truncated so that it has a predetermined number of levels. In nHDP, each node also has a predetermined number of children, and nCRP uses hyperparameters to control the number. As such, the two methods in effect require the user to provide the structure of an hierarchy as input.

As mentioned in the introduction, HLTA models document collections without referring to a document generation process. Instead, it uses hierarchical latent tree models (HLTMs) and the latent variables in the models are regarded as unobserved attributes of the documents.

The concept of latent tree models was introduced in~\citep{zhang02hierarchical,zhang04hierarchical}, where they were referred to as hierarchical latent class models. The term ``latent tree models" first appeared in~\citep{zhang08latent,wang08alatent}. Latent tree models generalize two classes of models from the previous literature. The first class is
  latent class models \citep{lazarsfeld1968, knott1999latent}, which are used for categorical data clustering in social sciences and medicine. The second class is probabilistic phylogenetic trees \citep{durbin98bio}, which are a tool for determining the evolution history of a set of species.

The reader is referred to~\cite{mourad2013survey} for a survey of research activities on latent tree models. The activities take place in three settings.
   In the first setting, data are assumed to be generated from an unknown LTM,\footnote{Here data generated from a model are vectors of values for observed variables, not documents.} and
the task is to recover the generative model \citep{choi11learning}. Here one tries to discover relationships between the latent structure and observed marginals that
hold in LTMs, and then use those relationships to reconstruct the true latent
structure from data. And one can prove theoretical results on consistency and sample complexity.

In the second setting, no assumption is made about how data are
generated and the task is to fit an LTM to data \cite{chen2012model}. Here it does not make sense to talk about theoretical guarantees on consistency and sample complexity. Instead, algorithms are evaluated empirically using held-out likelihood.
It has been shown that, on real-world datasets, better models can be obtained using methods developed in this setting than using those developed in the first setting \cite{liu2013greedy}. The reason is that, although the assumption of the first setting is reasonable for data from domains such as phylogeny, it is not reasonable for other types of data such as text data and survey data.

The third setting is similar to the second setting, except that model fit
is no longer the only concern. In addition, one needs to consider how useful
the results are to users, and might want to, for example, obtain a hierarchy
of latent variables.
 Liu {\em et al.}\ \citep{liu2014hierarchical} are the first to use latent tree models for hierarchical topic detection. They propose an algorithm, namely HLTA, for learning HLTMs from text data and give a method for extracting topic hierarchies from the models.
A method for scaling up the algorithm is proposed by Chen {\em et al.}\ \citep{chen2016progressive}.
This paper is based on
\citep{liu2014hierarchical,chen2016progressive}\footnote{NOTE TO REVIEWER (to be removed in the final version): Those are conference papers by the authors themselves. It is stated in the AIJ review form that ``a paper is novel if the results it describes
were not previously published {\bf by other authors}, and were not previously published
 by the same authors {\bf in any archival journal}".}. There are substantial extensions: The novelty of  HLTA w.r.t the LDA-based methods is now systematically discussed; The theory and algorithm are described in more details and two practical issues are discussed; A new parameter estimation method is  used for large datasets; And the empirical evaluations are more extensive.

Another method to learn a hierarchy of latent variables from data is proposed by Ver Steeg and Galstyan \cite{versteeg14discovering}. The method is named {\em correlation explanation (CorEx)}. Unlike HLTA, CorEx is proposed as a model-free method and it hence does not intend to provide a representation for the joint distribution of the observed variables.
% and its output can hence be called a {\em CorEx tree}. In both CorEx trees and HLTMs, a latent variable $Y$ is introduced to explain the correlations among a set of other variables
%
%
% in the way the relationship between a latent variable and its children is quantified. In HLTMs, the relationship is characterized using conditional distributions of the children given the latent variable, and there is a separate distribution for each child. The total number of parameters is linear in the number of children. In CorEx trees, on the other hand, the relationship is characterized by a conditional distribution of the latent variable given the children. The total number of parameters is exponential in the number of children.

% CorEx trees are obtained by recursively optimizing an information-theoretic criterion, once at each level. HLTMs are obtained by recursively fitting flat latent tree models to data, once at each level. Due to its similarity to HLTA, CorEx is included in our empirical evaluations, along with PAM, nCRP and nHDP.

HLTA produces a hierarchy with word variables at the bottom and multiple levels of latent variables at top. It is hence related to hierarchical variable clustering. One difference is that HLTA also partitions documents while variable clustering does  not. There is a vast literature on document clustering \cite{steinbach2000comparison}. In particular, co-clustering \cite{dhillon2001co} can identify document clusters where each cluster is associated with a potentially different set of words. However, document clustering and topic detection are generally considered two different fields with little overlap. This paper bridges the two fields by developing a full-fledged HTD method that partitions documents in multiple ways.

\section{Latent Tree Models}
\label{sec.ltm}
A {\em latent tree model (LTM)} is a tree-structured Bayesian network \citep{pear1988probabilistic}, where the leaf nodes represent observed variables and the internal nodes represent latent variables. An example is shown in Figure 1 (a). In this paper, all variables are assumed to be binary. The model parameters include a marginal distribution for the root $Z_1$, and a conditional distribution for each of the other nodes given its parent. The product of the distributions defines a joint distribution over all the variables.

\begin{figure}[t]
\includegraphics[width=12cm]{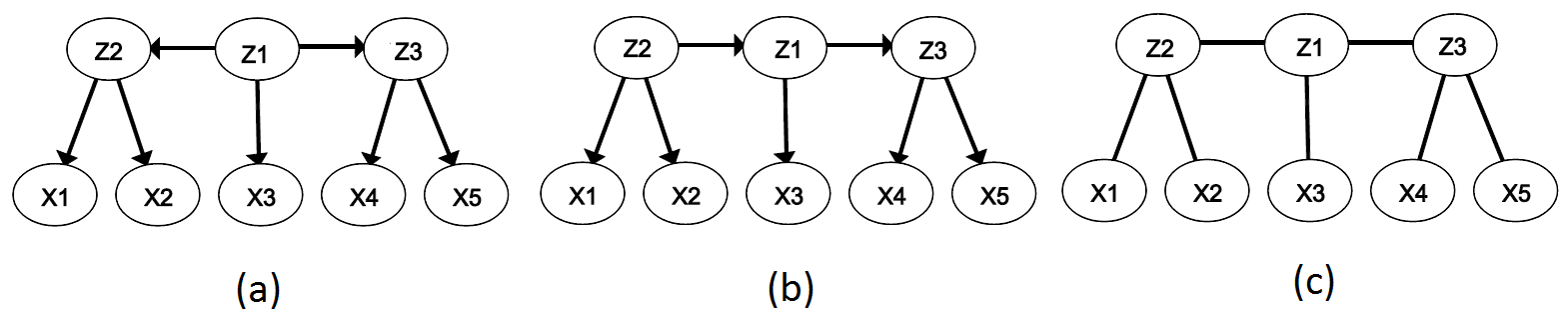}
\caption{\small The undirected latent tree model in (c) represents an equivalent class of directed latent tree models, which includes (a) and (b) as members.}
\label{fig.ltm}
\end{figure}

In general,  an LTM has $n$ observed variables ${\bf X}=\{X_1,\ldots,X_n\}$ and $m$ latent variables ${\bf Z}=\{Z_1,\ldots,Z_m\}$. Denote the parent of a variable $Y$ as $pa(Y)$ and let $pa(Y)$ be a empty set when $Y$ is the root. Then the LTM defines a joint distribution over all observed  and latent variables as follows:
\begin{eqnarray}
{P(X_1,\ldots,X_n, Z_1,\ldots,Z_m)} = \prod_{Y \in {\bf X}\cup {\bf Z}}P(Y \mid pa(Y))
\end{eqnarray}

By changing the root from $Z_1$ to $Z_2$ in Figure 1 (a), we get another model shown in  (b). The two models are {\em equivalent} in the sense that they represent the same set of distributions over the observed variables $X_1$, \ldots, $X_5$ \citep{zhang04hierarchical}. It is not possible to distinguish between equivalent models based on data. This implies that the root of an LTM, and hence  orientations of edges, are unidentifiable. It therefore makes more sense to talk about undirected LTMs, which is what we do in this paper.  One example is shown in Figure 1 (c). It represents an equivalent class of directed models. A member of the class can be obtained by picking a latent node as the root and directing the edges away from the root. For example,  (a) and (b) are obtained from (c) by choosing $Z_1$  and $Z_2$ to be the root respectively.  In implementation, an undirected model is represented using an arbitrary directed model in the equivalence class it represents.

In the literature, there are variations of LTMs where some internal nodes are observed \citep{choi11learning} and/or the variables are continuous \citep{poon2010variable,kirshner2012latent,song2014nonparametric}. In this paper, we focus on basic LTMs as defined in the previous two paragraphs.

We use $|Z|$ to denote  the number of possible states of a variable $Z$.
An LTM is {\em regular} if, for any latent node $Z$, we have that
\begin{eqnarray}
|Z| \leq \frac{\prod_{i=1}^k|Z_i|}{\max_{i=1}^k|Z_i|},
\end{eqnarray}
\noindent where $Z_1$, \ldots, $Z_k$ are the neighbors of $Z$, and that the inequality holds strictly when $k=2$. When all variables are binary, the condition reduces to that each latent node must have at least three neighbors.

 For any irregular LTM, there is a regular model that has fewer parameters and represents that same set of distributions over the observed variables \citep{zhang04hierarchical}. Consequently, we focus only on regular models.

\section{Hierarchical Latent Tree Models and Topic Detection}
\label{sec.hltm}

We will later present an algorithm, called HLTA,  for learning from text data models such as the one shown in Figure \ref{fig.toy-hltm}. There is a layer of observed variables at the bottom and multiple layers  of  latent variables on top. The model is hence called a \emph{hierarchical latent tree model (HLTM)}.
In this section, we discuss how to interpret HLTMs and how to extract topics and topic hierarchies from them.

\subsection{HLTMs for Text Data}
\begin{figure*}[t]
\begin{center}
\includegraphics[width=12cm]{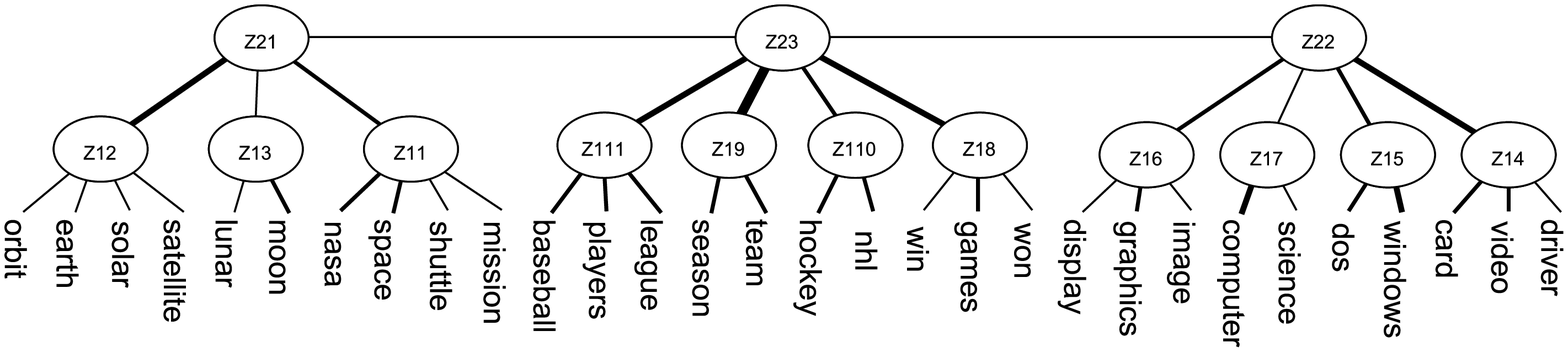}

\caption{Hierarchical latent tree model obtained from a toy text dataset.  The latent variables right above the word variables represent word co-occurrence patterns and those at higher levels represent co-occurrence of patterns at the level below.}
\label{fig.toy-hltm}
\end{center}
%\vspace{\floatingsep}
\end{figure*}

We use the toy model in Figure~\ref{fig.toy-hltm} as an running example. It is learned from a subset of the 20 Newsgroup data\footnote[1]{http://qwone.com/~jason/20Newsgroups/}. The variables at the bottom level, level 0, are observed binary variables that represent the presence/absence of words in a document. The latent variables at level 1 are introduced during data analysis to model word co-occurrence patterns. For example, $Z11$ captures the probabilistic co-occurrence of the words \word{nasa}, \word{space}, \word{shuttle} and \word{mission}; $Z12$ captures the probabilistic co-occurrence of the words \word{orbit}, \word{earth}, \word{solar} and \word{satellite}; $Z13$ captures the probabilistic co-occurrence of the words \word{lunar} and \word{moon}.
Latent variables at level 2 are introduced during data analysis to model the co-occurrence of the patterns at level 1. For example, $Z21$ represents the probabilistic co-occurrence of the patterns $Z11$, $Z12$ and $Z13$.

  Because the latent variables are introduced layer by layer, and each latent variable is introduced to explain the correlations among a group of variables at the level below, we regard, for the purpose of model interpretation, the edges between two layers as directed  and they are directed downwards. (The edges between top-level latent variables are not directed.) This allows us to talk a about the subtree rooted at a latent node. For example, the subtree rooted at $Z_{21}$ consists of the observed variables \word{orbit}, \word{earth}, \ldots, \word{mission}.

\subsection{Topics from HLTMs}

 There are totally 14 latent variables in the toy example. Each latent variable has two states and hence partitions the document collection into two soft clusters. To figure out what the partition and  the two clusters are about, we need to consider the relationship between the latent variable and the observed variables in its subtree. Take $Z21$ as an example. Denote the two document clusters it gives   as $Z21=s0$ and $Z{21}=s1$. The occurrence probabilities    in the two clusters of the words in the $Z21$ subtree is given in Table~\ref{table.z21}, along with the sizes of the two clusters.
 We see that the cluster $Z21$=s1 consists of $5\%$ of the documents. In this cluster, the words such as \word{space}, \word{nasa} and \word{orbit} occur with relatively high probabilities. It is clearly a meaningful and is interpreted as a {\em topic}. One might label the topic ``NASA''. The other cluster $Z21$=s0 consists of $95\%$ of the documents. In this cluster, the words occur with low probabilities. We interpret it as a {\it background topic}.

\begin{table}[t]
\caption{Document partition given by latent variable $Z21$.}
\label{table.z21}
{\small
\begin{center}
\begin{tabular}{|l|c|c|} \hline
$Z21$ & s0 (0.95) & s1 (0.05) \\ \hline
\word{space} & 0.04 & 0.58 \\
\word{nasa} & 0.03 & 0.43\\
\word{orbit} & 0.01& 0.33 \\
\word{earth} & 0.01 & 0.33 \\
\word{shuttle} & 0.01 & 0.24 \\
\word{moon}& 0.02 &0.26\\
\word{mission}&0.01&0.21\\ \hline
\end{tabular}
\end{center}
}
\end{table}

There are three subtle issues concerning Table~\ref{table.z21}. The first issue is how the word variables are ordered.
To answer the question, we need
   the {\em mutual information} (MI) $I(X; Y)$ \cite{cover06elements} between the two  discrete variables $X$ and $Y$, which is defined as follows:
 \begin{eqnarray}
I(X; Y) = \sum_{X, Y}P(X, Y) \log \frac{P(X, Y)}{P(X)P(Y)},
\label{eq.mi}
\end{eqnarray}
\noindent
In Table~\ref{table.z21}, the word variables are ordered according to their  mutual information with $Z21$.
 The words placed at the top of the table have the highest MI with $Z21$. They are the best ones to characterize the difference between the two clusters because their occurrence probabilities in the two clusters differ the most. They occur with high probabilities in the clusters $Z21=s1$ and with low probabilities in $Z21=s0$. If one is to choose only the top, say 5, words to characterize the topic $Z21$=s1, then the best words to pick are \word{space}, \word{nasa}, \word{orbit}, \word{earth} and \word{shuttle}.

The second issue is how  the background topic is determined. The answer is that, among the two document clusters given by $Z21$, the one where the words occur with lower probabilities is regarded as the background topic. In general, we consider the sum of the probabilities of the top 3 words. The cluster where the sum is lower is designated to be the background topic and labeled $s0$, and the other one is considered a genuine topic and labeled  $s1$.

Finally, the creation of Table~\ref{table.z21} requires the joint distribution of $Z21$ with each of the words variable in its subtrees (e.g., $P$(\word{space}, $Z21$). The distributions can be computed using Belief Propagation \citep{pear1988probabilistic}. The computation takes linear time  because the model is tree-structured.

\subsection{Topic Hierarchies from HLTMs}

If the background topics are ignored, each latent variable gives us exactly one topic.  As such, the model in Figure~\ref{fig.toy-hltm} gives us 14 topics, which are shown in Table~\ref{table.toy-topics}.  Latent variables at high levels of the hierarchy capture long-range word co-occurrence patterns and hence give thematically more general topics, while those at low levels of the hierarchy capture short-range word co-occurrence patterns and give thematically more specific topics. For example, the topic given by $Z22$ (\word{windows}, \word{card}, \word{graphics}, \word{video}, \word{dos}) consists of a mixture of words about several aspects of computers. We can say that the topic is about computers. The subtopics are each concerned with only one aspect of computers: $Z14$ (\word{card}, \word{video}, \word{driver}), $Z15$ (\word{dos}, \word{windows}), and $Z16$ (\word{graphics}, \word{display}, \word{image}).

\begin{table}[t]
\caption{Topic hierarchy given by the model in Figure~\ref{fig.toy-hltm}.}
\label{table.toy-topics}
\begin{small}
{\tt \scriptsize
\centering
\begin{tabular}{l}
{\bf  [0.05] space nasa orbit earth shuttle }\\
\quad [0.06] orbit earth solar satellite \\
\quad [0.05] space nasa shuttle mission \\
\quad [0.03] moon lunar \\ \\

{\bf [0.14] team games players season hockey}\\

\quad [0.14] team season \\
\quad [0.11] players baseball league \\
\quad [0.09] games win won \\
\quad [0.08] hockey nhl \\
\\

{\bf [0.24] windows card graphics video dos} \\

\quad [0.12] card video driver \\
\quad [0.15] windows dos \\
\quad [0.10] graphics display image \\
\quad [0.09] computer science \\

\end{tabular}

}
\end{small}
\end{table}

%An HLTM consists of a hierarchy of latent variables, each giving a topic. So, there is a hierarchy of topics.  The topic hierarchy given by the model of Figure~\ref{fig.toy-hltm} is shown in Table~\ref{table.toy-topics}. Each latent variable represents a partition of documents based on different co-occurrence patterns as shown by $Z23$. The latent variables at the lowest level, which we call level-1 latent variables, capture very small word co-occurrence patterns.   However, as for topic detection, level-1 word co-occurrence patterns may be specific. Higher level latent variables are hence introduced to identify co-occurrence patterns of the lower level variables,  giving more general topics

\subsection{More on Novelty}

In the introduction, we have discussed three differences between HLTA and the LDA-based methods. There are three other important differences. The fourth difference lies in the relationship between topics and documents. In the
LDA-based methods, a document is a mixture of topics, and the probabilities of the topics within a document sum to 1. Because of this, the LDA models are sometimes called {\em mixed-membership models}. In HLTA, a topic is a soft cluster of documents, and a document might belong to multiple topics with probability 1. In this sense, HLTMs can be said to be {\em multi-membership models}.

The fifth difference between HLTA and the LDA-based methods is about the semantics of the hierarchies they produce. In the context of document analysis, a common concept of hierarchy is a rooted tree where each node represents a cluster of documents, and the cluster of documents at a node is the union of the document clusters at its children. Neither HLTA nor the LDA-based methods yield such hierarchies. nCRP and nHDP produce a tree of topics. The topics at higher levels appear more often than those at lower levels, but they are not necessarily related thematically. PAM yields a collection of topics that are organized into a directed acyclic graph. The topics at the lowest level are distributions over words, and topics at higher levels are distributions over topics at the level below and hence are called super-topics.
In contrast, the output of HLTA is a tree of latent variables. Latent variables at high levels of the tree
capture long-range word co-occurrence patterns and hence give thematically more
general  topics, while latent variables at low levels of the tree capture
short-range word co-occurrence patterns and hence give thematically more specific
 topics.

 Finally, LDA-based methods require the user to provide the structure of a hierarchy, including the number of latent levels and the number of nodes at each level.
The number of latent levels is usually set at 3 out of efficiency considerations. The contents of the nodes (distributions over vocabulary) are learned from data. In contrast,
 HLTA learns both model structures and model parameters from data.
 The number of latent levels is not limited to 3.

\section{Model Structure Construction}

\label{sec.modelconstruct}

%In this section, we describe our new method PEM-HLTA which learns a hierarchical latent tree model for topic detection. Conceptually, the method proceeds in three steps: (1) Discover patterns of word co-occurrences; (2) Build a hierarchy by recursively discovering co-occurrence patterns of patterns; and (3) Extract topics from the resulting hierarchy. The pseudo code of the algorithm is given in Algorithm~\ref{alg.top}. We have explained the part (3) in Section~\ref{sec:TopicExtraction}. Now we first give the top-level control over PEM-HLTA and then describe the first two steps in details in the following subsections.
We present the HLTA algorithm in this and the next two sections. The inputs to HLTA include a collection of documents and several algorithmic parameters. The outputs include an HLTM and a topic hierarchy extracted from the HLTM. Topic hierarchy extraction has already been explained in Section~\ref{sec.hltm}, and we will hence focus on how to learn the HLTM. In this section we will describe the procedures for constructing the model structure. In Section \ref{sec.paramest} we will discuss parameter estimation issues, and Section \ref{sec.big} we discuss techniques employed to accelerate the algorithm.

\subsection{Top-Level Control}
\label{sec.topcontrol}
\begin{algorithm}[t]
\begin{description}
\item[Inputs:] $\mathcal{D}$ --- a collection of documents, $\tau$ --- upper bound on the number of top-level topics, $\mu$ --- upper bound on island size, $\delta$ --- threshold used in UD-test, $\kappa$ --- number of EM steps on final model. \\
\item[Outputs:] An HLTM and a topic hierarchy.
\end{description}
\small
\begin{algorithmic}[1]
\State $\mathcal{D}_1 \gets \mathcal{D}$, $m \gets null$.
\Repeat \label{alg1.start}
    \State $m_1 \gets$  \sysname{LearnFlatModel}($\mathcal{D}_1$, $\delta$, $\mu$);\label{alg1:learnFlat}

    \If{${m}= null$}
        \State ${m} \gets {m}_1$;\label{alg1.m}
    \Else
        \State $m\gets$ \sysname{StackModels}(${m}_1$, ${m}$); \label{alg1.stackmodel}
    \EndIf
    \State $\mathcal{D}_1 \gets$ \sysname{HardAssignment}($m$, $\mathcal{D}$); \label{alg1.hardassign}

\Until number of top-level nodes in $m  \leq \tau$.  \label{alg1.end}
\State Run EM on $m$ for $\kappa$ steps. \label{alg1.em}
\State \Return $m$ and topic hierarchy extracted from $m$. \label{alg1.return}
\end{algorithmic}
%\end{multicols}
\caption{\sysname{HLTA}($\mathcal{D}$, $\tau$, $\mu$, $\delta$, $\kappa$)}
\label{alg.top}
\end{algorithm}

The top-level control of HLTA is given in Algorithm~\ref{alg.top} and the subroutines are given in Algorithm~\ref{alg.learnFlat}-\ref{alg.pem-ltm}. In this subsection, we illustrate the top-level control using the toy dataset mentioned in Section~\ref{sec.hltm}, which involves 30 word variables.

There are 5 steps.  The first step (line~\ref{alg1:learnFlat}) yields the model shown in Figure~\ref{fig.overall} (a). It is said to be a {\em flat LTM} because each latent variable is connected to at least one observed variable. In  hierarchical models such as the one shown in Figure~\ref{fig.toy-hltm}, on the other hand, only the latent variables at the lowest latent layer are connected to observed variables, and other latent variables are not.  The learning of a flat model is the key step of HLTA. We will discuss it in details later.

{
\begin{figure*}[t]
\begin{center}
\includegraphics[width=12cm]{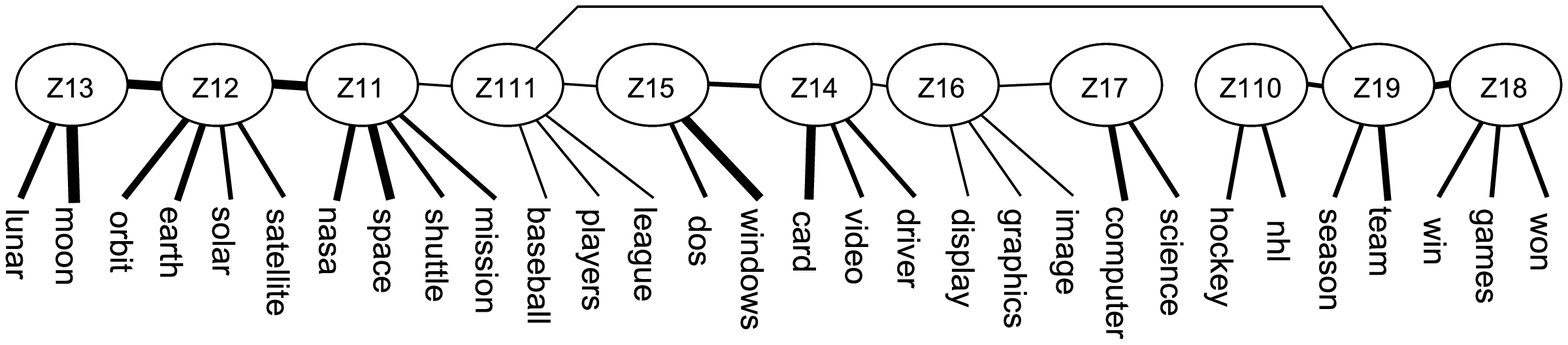} \\ (a) \\

\includegraphics[width=12cm]{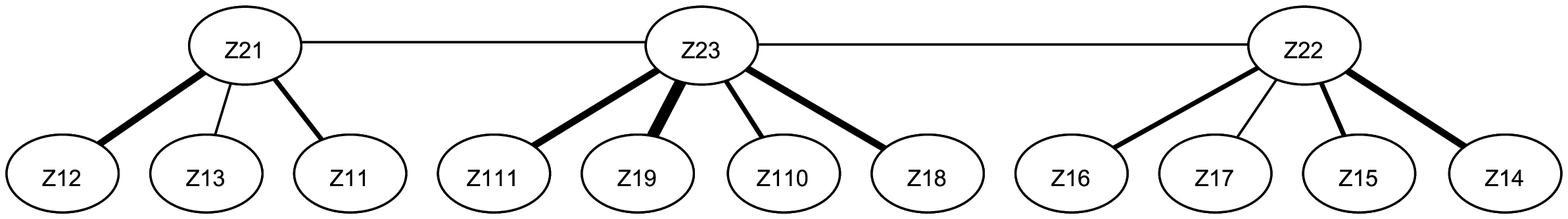} \\  (b)
\end{center}
\vspace{-2mm}
\caption{ Illustration of the top-level control of HLTA: (a) A flat model over the word variables is first learned; (b) The latent variables in (a) are converted into observed variables through data completion and another flat model is learned for them; Finally, the flat model in (b) is stacked on top of the flat model in (c) to obtain the hierarchical model in Figure~\ref{fig.toy-hltm}.}
\label{fig.overall}
\end{figure*}
}

We refer to the latent variables in the flat model from the first step as level-1 latent variables.
The second step (line~\ref{alg1.hardassign}) is to turn the level-1 latent variables into observed variables through data completion. To do so, the subroutine \sysname{HardAssignment} carries out inference to compute the posterior distribution of each latent variable for each document. The document is assigned to the state with the highest posterior probability, resulting in a dataset $\mathcal{D}_1$ over the level-1 latent variables.

In the third step, line~\ref{alg1:learnFlat} is executed again to learn a flat LTM  for the level-1 latent variables, resulting the model  shown in
Figure~\ref{fig.overall} (b).

 In the fourth step (line~\ref{alg1.stackmodel}), the flat model for the level-1 latent variables is stacked on top of the flat model for the observed variables, resulting
 in the hierarchical model in Figure~\ref{fig.toy-hltm}. While doing so, the subroutine \sysname{StackModels} cuts off the links among the level-1 latent variables.
 The parameter values for the new model are copied from two source models.

In general, the first four steps are repeated  until the number of top-level latent variables falls below a user-specified upper bound  $\tau$ (lines~\ref{alg1.start} to~\ref{alg1.end}). In our running example, we set $\tau = 5$. The number of nodes at the top level in our current model is 3, which is below the threshold $\tau$. Hence the loop is exited.

In the fifth step (line~\ref{alg1.em}), the EM algorithm~\cite{dempster77} is run on the final hierarchical model for $\kappa$ steps to improve its parameters, where $\kappa$  is another user specified input parameter.

The five steps can be grouped into two phases conceptually. The {\em model construction phase} consists of the first four steps. The objective is to build a hierarchical model structure.  The {\em parameter estimation phase} consists of the fifth step. The objective is to optimize the parameters of the hierarchical structure from the first phase.

\subsection{Learning Flat Models}
\label{sec.learnFlat}

The objective of  the flat model learning step is to find, among all flat models, the one that have the highest BIC score.   The BIC score \cite{Schwarz1978} of a model $m$ given a dataset $\mathcal{D}$  is defined as follows:
\begin{eqnarray}
\mathcal{BIC}(m\mid \mathcal{D}) = \log P(\mathcal{D}\mid m, \theta^*) - \frac{d}{2} \log |\mathcal{D}|,
\end{eqnarray}
where $\theta^*$ is the maximum likelihood estimate of the model parameters, $d$ is the number of free model parameters, and $|\mathcal{D}|$ is the sample size. Maximizing the BIC score intuitively means to find a model that fits the data well and that is not overly complex.

{
\begin{figure*}[t]
\begin{center}
\includegraphics[width=12cm]{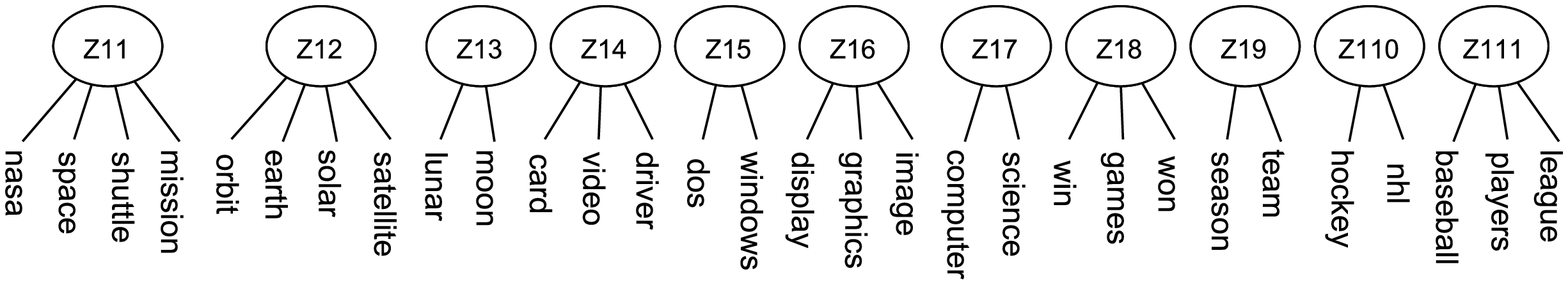}
\end{center}
\vspace{-2mm}
\caption{The subroutine \sysname{BuildIslands} partitions word variables into uni-dimensional
clusters and introduce a latent variable to each cluster to form an island (an LCM).}
\label{fig.islands}
\end{figure*}
}

\begin{algorithm}[t]
\small
\begin{algorithmic}[1]
 \State $\mathcal{L} \gets$ \sysname{BuildIslands}($\mathcal{D}$, $\delta$, $\mu$);\label{alg1.buildisland}
    \State $m \gets$  \sysname{BridgeIslands}($\mathcal{L}$, $\mathcal{D}_1$);\label{alg1:bridge}
    \State \Return $m$.
\end{algorithmic}
\caption{\sysname{LearnFlatModel}($\mathcal{D}$, $\delta$, $\mu$)}
\label{alg.learnFlat}
\end{algorithm}

One way to solve the problem is through search. The state-of-the-art in this direction is an algorithm named EAST~\cite{chen2012model}. It has been shown~\cite{liu2013greedy} to find better models that alternative algorithms such as BIN~\cite{harmeling11greedy} and CLRG~\cite{choi11learning}. However, it does not scale up. It is capable of handling data with only dozens of observed variables and is hence not suitable for text analysis.

In the following, we present an algorithm that, when combined with the parameter estimation technique to be described in the next section, is efficient enough to deal with large text data. The pseudo code
is given in Algorithm \ref{alg.learnFlat}. It calls two subroutines.
 The first subroutine  is \sysname{BuildIslands}. It partitions all word variables into clusters, such that the words in each cluster tend to co-occur and the co-occurrences can be properly modeled using a single latent variable. It then introduces a latent variable for each cluster to model the co-occurrence of the words inside it. In this way for each cluster we obtain an LTM with a single latent variable, and  is called a {\em latent class model (LCM)}. In our running example, the results are shown in Figure~\ref{fig.islands}.
   We metaphorically refer to the LCMs as {\em islands}.

The second subroutine is \sysname{BridgeIslands}. It  links up the islands by first estimating the  mutual information  between every pair of latent variables, and then finding the maximum spanning tree~\cite{ChowL68}.  The result is the model in Figure~\ref{fig.overall} (a).

We now set out to describe the two subroutines in details.

\subsubsection{Uni-Dimensionality Test}
\label{sec.udtest}
Conceptually, a set of variables is said to be {\it uni-dimensional} if the correlations among them can be properly modeled using a single latent variable. Operationally, we rely on the {\it uni-dimensionality test (UD-test)} to determine whether a set of variables is uni-dimensional.

%\begin{figure}[t]
%\begin{center}
%  \begin{subfigure}[b]{0.35\linewidth}
%    \centering
%  \includegraphics[scale=.25]{figs/new1.pdf}
%    \caption{$m_1$}
%    \end{subfigure}
%    \begin{subfigure}[b]{0.35\linewidth}
%    \centering
%  \includegraphics[scale=.25]{figs/new2.pdf}
%    \caption{$m_2$}
%    \end{subfigure}
%\caption{\small  The two latent tree models that are used in the UD-test.}
%\label{fig.ud-test}
%\end{center}
%\end{figure}

To perform UD-test on a set $\mathcal{S}$ of observed variables, we first learn two latent tree models $m_1$ and $m_2$ for $\mathcal{S}$ and then compare their BIC scores. The model $m_1$ is the model with the highest BIC score among all LTMs with a single latent variable, and the model $m_2$ is the model with the highest BIC score among all LTMs with two latent variables. Figure~\ref{fig.oneisland} (b) shows what the two models might look like when $\mathcal{S}$ consists of four word variables
\word{nasa}, \word{space},\word{shuttle} and \word{mission}. We conclude that $\mathcal{S}$ is uni-dimensional if the following inequality holds:
\begin{eqnarray}
\mathcal{BIC}(m_2\mid \mathcal{D}) - \mathcal{BIC}(m_1\mid\mathcal{D}) < \delta,
\label{eq.ud-test}
\end{eqnarray}
where $\delta$ is a user-specified threshold.  In other words, $\mathcal{S}$ is considered uni-dimensional if the best two-latent variable model is not significantly better than the best one-latent variable model.

Note that the UD-test is related to the Bayes factor for comparing the two models~\cite{raftery95bayesian}:
\begin{eqnarray}
K=\dfrac{P(\mathcal{D}|m_2)}{P(\mathcal{D}|m_1)}.
\end{eqnarray}
The strength of evidence in favor of $m_2$ depends on the value of $K$. The following guidelines are suggested in~\cite{raftery95bayesian}:  If the quantity ~$2 \log K$~ is from 0 to 2, the evidence is negligible; If it  is between 2 and 6, there is positive evidence in favor of $m_2$;  If it is between 6 to 10, there is strong evidence in favor of $m_2$; And if it is larger than 10, then there is very strong evidence in favor of $m_2$.  Here, ``$\log$" stands for natural logarithm.

It is well known that the BIC score $\mathcal{BIC}(m| \mathcal{D})$ is a large sample approximation of the marginal loglikelihood $\log P(\mathcal{D}|m)$~\cite{Schwarz1978}. Consequently, the difference $\mathcal{BIC}(m_2|\mathcal{D}) - \mathcal{BIC}(m_1|\mathcal{D})$ is  a large approximation of the logarithm of the Bayes factor $\log K$.  According to the cut-off values for the Bayes factor, we conclude that there is {\it positive, strong, and very strong} evidence favoring $m_2$ when the difference is larger than 1, 3 and 5 respectively. In our experiments, we always set $\delta=3$.

%The UD-test is time-consuming especially when the variables set $\mathcal{S} $ and data size $ |\mathcal{D}|$ grow larger. We take three measures to make it efficient enough for large datasets: (1) Restrict the latent variables in $m_1$ and $m_2$ to be binary, (2) heuristically pick a two-latent variable model for $m_2$ instead of searching for the best one, and (3) use progressive EM instead of EM for parameter estimation.
\begin{algorithm}[t]
\small
\begin{algorithmic}[1]
\State $\mathcal{V} \gets$ variables in $\mathcal{D}$, $\mathcal{L} \gets \emptyset$.
\While {$|\mathcal{V}| > 0$}
    \State $m \gets$ \sysname{OneIsland}($\mathcal{D}$, $\mathcal{V}$, $\delta$, $\mu$);
    \State $\mathcal{L} \gets \mathcal{L} \cup \{m\}$;
    \State $\mathcal{V} \gets $ variables in $\mathcal{D}$ but not in any $m \in \mathcal{L}$;
\EndWhile
\State \Return $\mathcal{L}$.
\end{algorithmic}
\caption{\sysname{BuildIslands}($\mathcal{D}$, $\delta$, $\mu$)}
\label{alg.buildislands}
\end{algorithm}

\begin{algorithm}[t]
\small
%\begin{multicols}{2}
\begin{algorithmic}[1]
\State \textbf{if} $|\mathcal{V}| \leq 3$, $m \gets$ \sysname{LearnLCM}($\mathcal{D}$, $\mathcal{V}$), \textbf{return} $m$.\label{alg3.line1}
\State $\mathcal{S} \gets$ two variables in $\mathcal{V}$ with highest MI; \label{alg3.s1}
\State {\scriptsize $X \gets \argmax_{A \in \mathcal{V} \setminus \mathcal{S}} MI(A, \mathcal{S})$};\label{alg3.s2}
\State {$\mathcal{S} \gets \mathcal{S} \cup {X} $};\label{alg3.s3}
\State $\mathcal{V}_1 \gets  \mathcal{V} \setminus \mathcal{S}$;
\State $\mathcal{D}_1 \gets$ \sysname{ProjectData}($\mathcal{D}$, $\mathcal{S}$); \\
$m \gets$ \sysname{LearnLCM}($\mathcal{D}_1$, $\mathcal{S}$).\label{alg3.firstlcm}
\Loop
    \State {\scriptsize $X \gets \argmax_{A \in \mathcal{V}_1} MI(A, \mathcal{S})$; \label{alg3.pick-x}
    \State $W \gets \argmax_{A \in S} MI(A, X)$}\label{alg3.pick-w};
    \State $\mathcal{D}_1 \gets$ \sysname{ProjectData}$(\mathcal{D}, \mathcal{S} \cup \{X\})$, $\mathcal{V}_1 \gets \mathcal{V}_1 \setminus \{X\}$;
    \State $m_1 \gets$ \sysname{Pem-Lcm}$(m, \mathcal{S}, X, \mathcal{D}_1)$;\label{alg3.lcm}
    \State \textbf{if} $|\mathcal{V}_1| = 0$ \Return $m_1$.\label{alg3.return-lcm}
    \State $m_2 \gets$ \sysname{Pem-Ltm-2l}($m$, $\mathcal{S} \setminus \{W\}$, $\{W, X\}$, $\mathcal{D}_1$);\label{alg3.ltm}
    \If{$BIC(m_2|\mathcal{D}_1) - BIC(m_1|\mathcal{D}_1) > \delta$}\label{alg3.compare}
        \State \Return $m_2$ with $W$, $X$ and their parent removed.\label{alg3.return-island}
    \EndIf
    \State \textbf{if} $|\mathcal{S}| \geq \mu$, \Return $m_1$.\label{alg3.maxisland}
    \State $m \gets m_1$, $\mathcal{S} \gets \mathcal{S} \cup \{X\}$.\label{alg3.add-x}
\EndLoop
\end{algorithmic}
%\end{multicols}
\caption{\sysname{OneIsland}($\mathcal{D}$, $\mathcal{V}$, $\delta$, $\mu$)}\label{alg.oneisland}
\end{algorithm}

\subsubsection{Building Islands}
\label{sec.buildislands}

The subroutine \sysname{Buildislands} (Algorithm~\ref{alg.buildislands}) builds islands one by one. It builds the first island by calling another subroutine \sysname{OneIsland} (Algorithm~\ref{alg.oneisland}). Then it removes the variables in the island from the dataset, and repeats the process to build other islands. It continues until all variables are grouped into islands.

The subroutine \sysname{OneIsland} (Algorithm~\ref{alg.oneisland}) requires a measurement of how closely correlated each pair of variables are. In this paper, mutual information is used for the purpose. The mutual information $I(X; Y)$  between the two variables $X$ and $Y$ is given by (\ref{eq.mi}).
 We will also need the mutual information (MI) between a variable $X$ and a set of variables $\mathcal{S}$. We estimate it as follows: \begin{eqnarray}
 {\text{MI}}(X, \mathcal{S})=\max_{A \in \mathcal{S}} {\text{MI}}(X, A).
 \end{eqnarray}

The subroutine \sysname{OneIsland}   maintains a working set $\mathcal{S}$ of observed variables.   Initially, $\mathcal{S}$ consists of the pair of variables with the highest MI (line~\ref{alg3.s1}), which will be referred to as the {\em seed variables} for the island. Then the variable that has the highest MI with those two variables is added to $\mathcal{S}$ as the third variable (line~\ref{alg3.s2} and~\ref{alg3.s3}).  Then other variables are added to $\mathcal{S}$ one by one. At each step, we pick the variable $X$ that has the highest MI with the current set $\mathcal{S}$ (line~\ref{alg3.pick-x}), and perform UD-test on the set $\mathcal{S} \cup \{X\}$ (lines~\ref{alg3.lcm}, \ref{alg3.ltm}, \ref{alg3.compare}). If the UD-test passes, $X$ is added to $\mathcal{S}$ (line~\ref{alg3.add-x}) and the process continues.  If the UD-test fails, one island is created and the subroutine returns (line~\ref{alg3.return-island}). The subroutine also returns when the size of the island reaches a user-specified upper-bound $\mu$ (line~\ref{alg3.maxisland}). In our experiments, we always set $\mu = 15$.

%$\mathcal{S} \setminus \{W\}$ is taken as a uni-dimensional subset of variables. We obtain an island for those variables by removing $W$, $X$ and their parent from $m_2$ (line~\ref{alg3.return-island}).  If $X$ is the last variable, the UD-test is skipped and $m_1$ is returned as the final island (line~\ref{alg3.return-lcm}).\footnote{When there are exactly two latent variables left, there is no need to introduce new latent variable for the two variables. Otherwise, it will cause irregular models~\cite{zhang04hierarchical}.}

%In the test, $m_1$ is an LCM over all the variables in $\mathcal{S} \cup \{X\}$, with parameters optimized by \sysname{PEM-LCM}.  Let $W$ be the variable in $\mathcal{S}$ that has the highest MI with $X$. For $m_2$ in the UD-test, we choose the model where one latent variable is connected to the variables in $\mathcal{S} \setminus \{W\}$ and the second latent variable connected to $W$ and $X$.
\begin{figure}[!htb]
\begin{center}
    \begin{subfigure}[b]{0.3\linewidth}
    \centering
  \includegraphics[scale=.15]{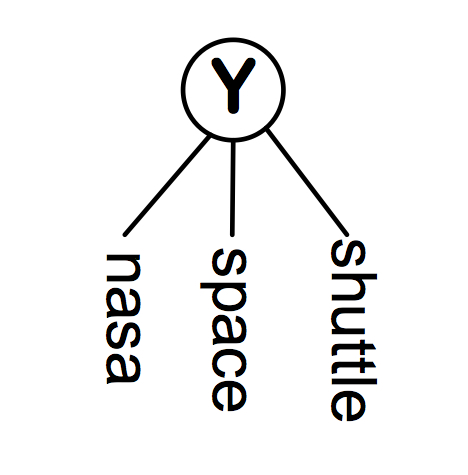}
   \vspace{-4mm}
    \caption{Initial island}
    \label{fig.oneislanda}
    \end{subfigure}
 \begin{subfigure}[b]{0.5\linewidth}
    \centering
  \includegraphics[scale=.15]{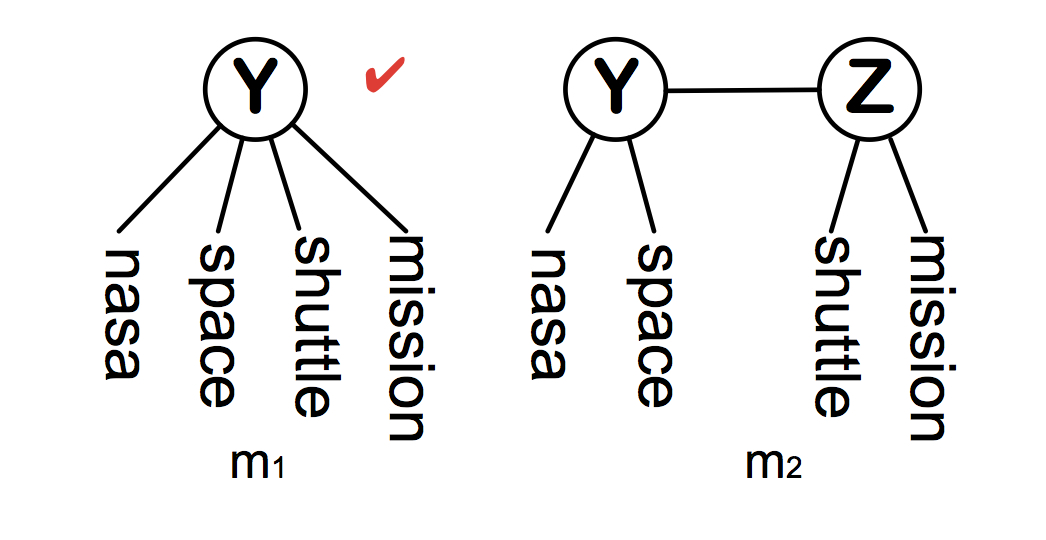}
      \vspace{-4mm}
    \caption{UD-test passes after adding \word{mission}}
    \label{fig.oneislandb}
    \end{subfigure}\\
       %\vspace{-2mm}
    \begin{subfigure}[b]{0.8\linewidth}
    \centering
  \includegraphics[scale=.15]{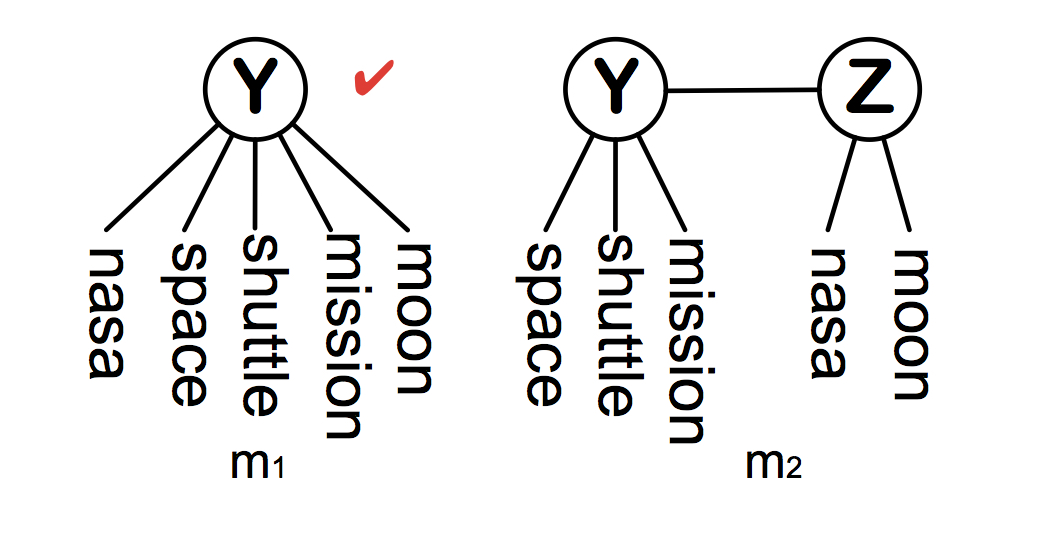}
     \vspace{-4mm}
      \caption{UD-test passes after adding \word{moon}}
      \label{fig.oneislandc}
    \end{subfigure}\\
       % \vspace{-2mm}
    \begin{subfigure}[b]{0.55\linewidth}
    \centering
  \includegraphics[scale=.15]{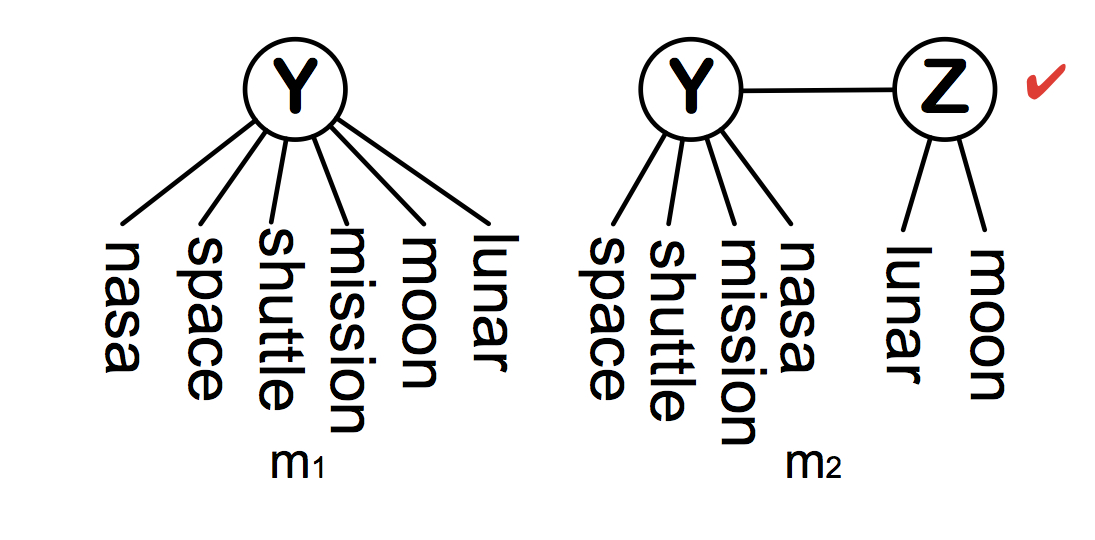}
      \vspace{-4mm}
  \caption{UD-test fails after adding \word{lunar}}
  \label{fig.oneislandd}
    \end{subfigure}
    \begin{subfigure}[b]{0.4\linewidth}
    \centering
  \includegraphics[scale=.15]{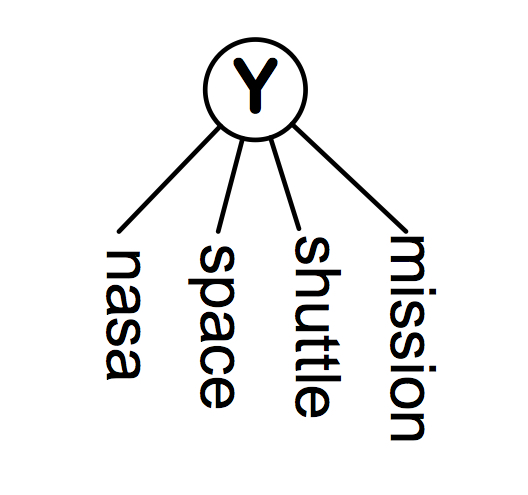}
     \vspace{-4mm}
  \caption{Final island}
  \label{fig.oneislande}
    \end{subfigure}
\caption{\small Illustration of the \sysname{OneIsland} subroutine.}
\label{fig.oneisland}
\end{center}
\end{figure}

The UD-test requires two models $m_1$ and $m_2$. In principle, they should be the best models with one and two latent variables respectively. For the sake of computational efficiency, we construct them heuristically in this paper. For $m_1$, we choose the LCM where the latent variable is binary and the parameters are optimized by a fast subroutine \sysname{Pem-Lcm} that will be described in the next section.

Let $W$ be the variable in $\mathcal{S}$ that has the highest MI with the variable $X$ to be added to the island. For $m_2$, we choose the model where one latent variable is connected to the variables in $\mathcal{S}\setminus \{W\}$ and the second latent variable connected to $W$ and $X$. Both latent variables are binary and the model parameters are optimized by a fast subroutine \sysname{Pem-Ltm-2l} that will be described in the next section.

Let us illustrate the \sysname{OneIsland} subroutine using an example in Figure~\ref{fig.oneisland}. The pair of variables \word{nasa} and \word{space} have the highest MI among all variables, and they are hence the {\em seed variables}. The variable \word{shuttle} has the highest MI with the pair among all other variables, and hence it is chosen as the third variable to start the island (Figure~\ref{fig.oneisland} (a)). Among all the other variables, \word{mission} has highest MI with the three variables in the model. To decide whether \word{mission} should be added to the group, the two models $m_1$ and $m_2$ in Figure~\ref{fig.oneisland} (b) are created. In $m_2$, \word{shuttle} is grouped with the new variable because it has the highest MI with the new variable among all the three variables in Figure~\ref{fig.oneisland} (a). It turns out that $m_1$ has higher BIC score than $m_2$. Hence the UD-test passes and the variable \word{mission} is added to the group. The next variable to be considered for addition is \word{moon} and it is added to the group because the UD-test passes again (Figure~\ref{fig.oneisland} (c)). After that, the variable \word{lunar} is considered. In this case, the BIC score of $m_2$ is significantly higher than that of $m_1$ and hence the UD-test fails (Figure~\ref{fig.oneisland} (d)). The subroutine \sysname{OneIsland} hence terminates. It returns an island, which is the part of the model $m_2$ that does not contain the last variable \word{lunar} (Figure~\ref{fig.oneisland} (e)). The island consists of the four words \word{nasa}, \word{space}, \word{shuttle} and \word{mission}. Intuitively, they are grouped together because they tend to co-occur in the dataset.

%As an illustration, suppose $\mathcal{S}$ initially consists of three variables $A$, $B$, $C$.  Let $D$ be the variable that has the maximum MI with $\mathcal{S}$ among all other variables. Suppose the UD-test passes on $\mathcal{S} \cup \{D\}$, then $D$ is added to $\mathcal{S}$.  Next let $E$ be the variable that has the maximum MI with $S$ among all other variables and $C$ be the variable in $\mathcal{S}$ that has the maximum MI with $E$. The UD-test is performed on $\mathcal{S} \cup \{E\}$.  The two models $m_1$ and $m_2$ used in the test are as shown in Figure~\ref{fig.ud-test}, where $Y$ and $Z$ are two latent variables.  If the test fails, then $C$, $E$ and $Z$ are removed from $m_2$, and what remains in the model, an LCM, is returned

%When the UD-test fails, model $m_2$ gives us two potential sibling clusters. If one of the two potential sibling clusters  contains both the two initial variables, it is picked. Otherwise,   BI picks the one with more variables and breaks ties arbitrarily. In the example, the two potential sibling clusters are $\{X_1, X_2, X_4\}$  and $\{X_3, X_5\}$. BI picks  $\{X_1, X_2, X_4\}$  because it contains both the two initial variables $X_1$ and $X_2$.

%After the first island is determined, \sysname{BuildIslands} removes the variables in the cluster from the dataset, and repeats the process to find other clusters. This continues until all variables are grouped into islands.

\subsubsection{Bridging Islands}
\label{sec.bridgeislands}

After the islands are created, the next step is to link them up so as to obtain a model over all the word variables. This is carried out by the \sysname{BridgeIslands} subroutine and the idea is borrowed from~\cite{chow68approximating}. The subroutine first estimates the MI between each pair of latent variables in the islands, then constructs a complete undirected graph with the MI values as edge weights, and finally finds the maximum spanning tree of the graph. The parameters of the newly added edges are estimated using a fast method that will be described at the end of  Section \ref{sec.pem-hlta}.

Let $m$ and $m'$ be two islands with latent variables $Y$ and $Y'$ respectively.  The MI $I (Y ; Y')$ between $Y$ and $Y'$  is calculated using Equation~(\ref{eq.mi}) from the following joint distribution:
\begin{eqnarray}
P(Y,Y'\mid \mathcal{D},m,m')=C\sum_{d\in \mathcal{D}}P(Y\mid m,d)P(Y'\mid m',d)
\label{eq.mi1}
\end{eqnarray}
where $P(Y\mid m, d)$ is the posterior distribution of $Y$ in $m$ given data case $d$ , $P(Y'\mid m', d)$ is that of $Y'$ in $m'$, and $C$ is the normalization constant.

In our running example, applying \sysname{BridgeIslands}  to the islands in Figure~\ref{fig.islands} results in the flat model shown in Figure~\ref{fig.overall} (a).\\

\section{Parameter Estimation during Model Construction}
 \label{sec.paramest}

In the model construction phase, a large number of intermediate models are generated. Whether HLTA can scale up depends on whether the parameters of those intermediate models and the final model can be estimated efficiently.
In this section, we present a fast method called \emph{progressive EM} for estimating the parameters of the intermediate models. In the next section, we will discuss how to estimate the parameters of the final model efficiently when the sample size is very large.

\subsection{The EM Algorithm}
\label{sec.EM}
We start by briefly reviewing the EM algorithm. Let ${\bf X}$ and ${\bf H}$ be respectively the sets of observed and latent variables in an LTM $m$, and let ${\bf V = X \cup H}$. Assume one latent variable is picked as the root and all edges are directed away from the root. For any $V$ in ${\bf V}$ that is not the root, the parent $\text{pa}(V)$ of $V$ is a latent variable and can take values `0' or `1'. For technical convenience, let $\text{pa}(V)$ be a dummy variable with only one possible value when $V$ is the root. Enumerate all the variables as $V_1, V_2, \cdots, V_n$.  We denote the parameters of $m$ as
\begin{eqnarray}
\label{eq.theta_ijk}
\theta_{ijk} = P(V_i=k|\text{pa}(V_i)=j),
\end{eqnarray}
where $i \in \{1, \cdots, n\}$, $k$ is value of $V_i$ and $j$ is a value of $\text{pa}(V_i)$ . Let $\theta$ be the vector of all the parameters.

Given a dataset $\mathcal{D}$, the loglikelihood function of $\theta$ is given by
\begin{eqnarray}
l(\theta \mid \mathcal{D}) = \sum_{d\in \mathcal{D}}\sum_{\bf H}\log P(d, {\bf H}\mid \theta).
\label{exp.ll}
\end{eqnarray}
\noindent The \emph{maximum likelihood estimate (MLE)} of $\theta$ is the value that maximizes the loglikelihood function.

Due to the presence of latent variables, it is intractable to directly maximize the loglikelihood function. An iterative method called the \emph{Expectation-Maximization (EM)}~\cite{dempster77} algorithm is usually used in practice. EM starts with an initial guess $\theta^{(0)}$ of the parameter values, and then produces a sequence of estimates $\theta^{(1)}$, $\theta^{(2)}$,$\cdots$. Given the current estimate $\theta^{(t)}$, the next estimate $\theta^{(t+1)}$ is obtained through an E-step and an M-step. In the context of latent tree models, the two steps are as follows:
\begin{itemize}
\item The E-step:
\begin{eqnarray}
\label{eq.EM-E}
n_{ijk}^{(t)} = \sum_{d \in \mathcal{D}} P(V_i=k, pa(V_i)=j|d, m, \theta^{(t)}).
\end{eqnarray}
\item The M-step:
\begin{eqnarray}
\label{eq.EM-M}
\theta_{ijk}^{(t+1)} = \frac{ n_{ijk}^{(t)}}{\sum_k n_{ijk}^{(t)}}.
\end{eqnarray}
\end{itemize}
Note that the E-step requires the calculation of  $P(V_i, pa(V_i)|d, m, \theta^{(t)})$ for each data case $d \in \mathcal{D}$ and each variable $V_i$. For a given data case $d$, we can calculate $P(V_i, pa(V_i)|d, m, \theta^{(t)})$ for all variables $V_i$ in linear time  using message propagation~\cite{murphy12machine}.

EM terminates when the improvements in loglikelihood  $l(\theta^{(t+1)}|\mathcal{D}) - l(\theta^{(t)}|\mathcal{D})$ falls below a predetermined threshold or when the number of iterations reaches a predetermined limit.  To avoid local maxima, multiple restarts are usually used.

\subsection{Progressive EM}
\label{sec.PEM}

\begin{figure}[t]
\begin{center}
  \begin{subfigure}[b]{0.45\linewidth}
    \centering
  \includegraphics[scale=.3]{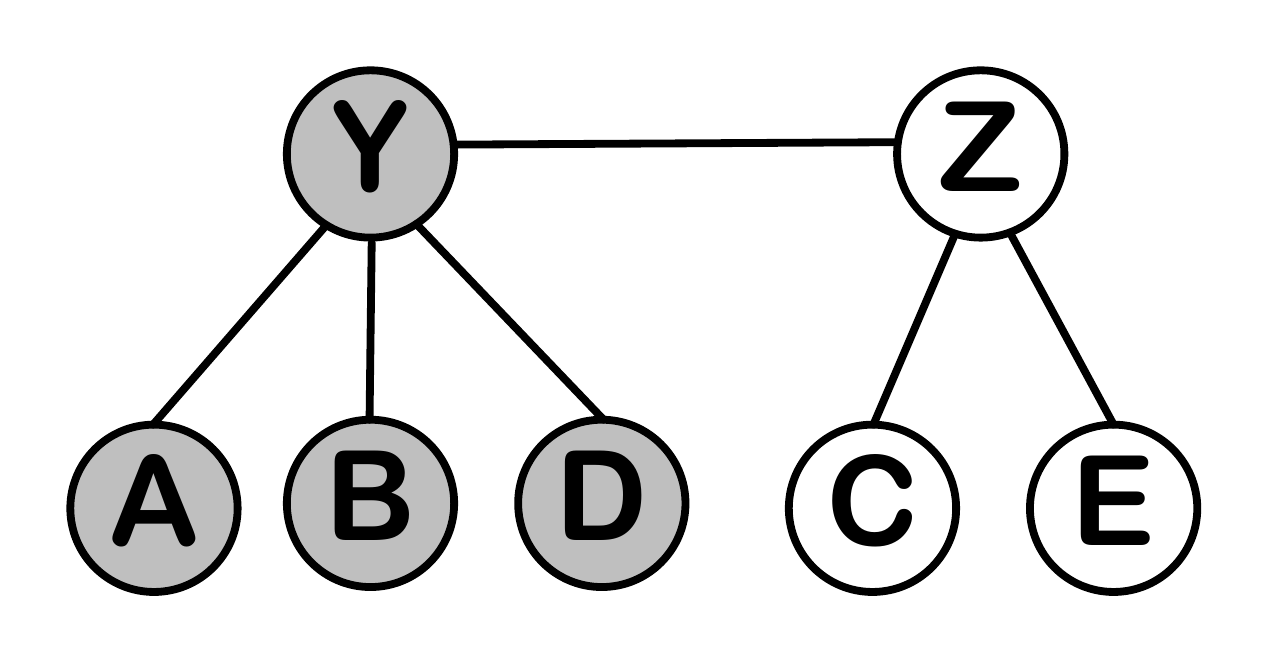}
    \caption{ }
    \label{fig.pema}
    \end{subfigure}
    \begin{subfigure}[b]{0.45\linewidth}
    \centering
  \includegraphics[scale=.3]{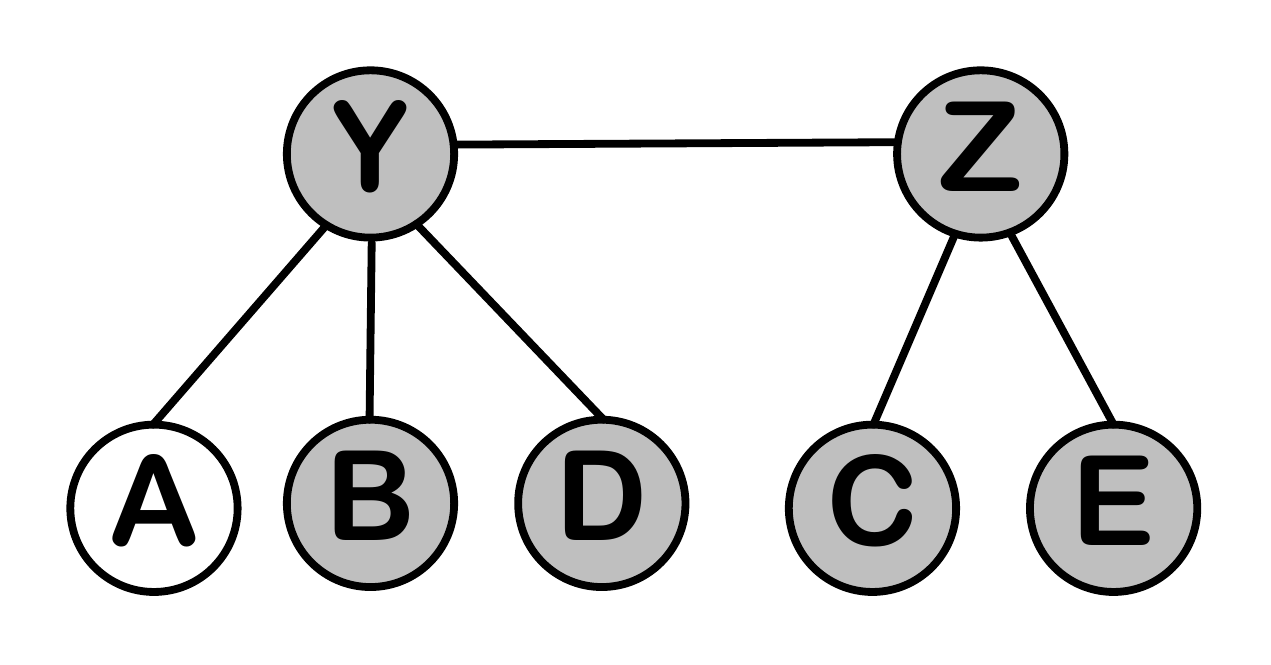}
    \caption{ }
    \label{fig.pemb}
    \end{subfigure}
\caption{\small Progressive EM: EM is first run in the submodel shaded in (a) to estimate the distributions $P(Y), P(A|Y), P(B|Y)$ and $P(D|Y)$, and then, EM is run in the submodel shaded in (b), with $P(Y)$, $P(B|Y)$ and $P(D|Y)$ fixed, to estimate the distributions $P(Z|Y), P(C|Z)$ and $P(E|Z)$.}
\label{fig.pempic}
\end{center}
\end{figure}

Being an iterative algorithm, EM can be trapped in local maxima. It is also time-consuming and does not scale up well. Progressive EM is proposed as a fast alternative to EM for the model construction phase.  It estimates all the parameters in multiple steps and, in each step, it considers a small part of the model and runs EM in the submodel to maximize the local likelihood function.  The idea is illustrated in Figure~\ref{fig.pempic}. Assume $Y$ is selected to be the root. To estimate all the parameters of the model, we first run EM in the part of the model shaded in Figure~\ref{fig.pema} to estimate $P(Y), P(A|Y), P(B|Y)$ and $P(D|Y)$, and then run EM in the part of the model shaded in Figure~\ref{fig.pemb}, with $P(Y), P(B|Y)$ and $P(D|Y)$ fixed, to estimate $P(Z|Y), P(C|Z)$ and $P(E|Z)$.

\subsection{Progressive EM and HLTA}
\label{sec.pem-hlta}
We use progressive EM to estimate the parameters for the intermediate models generated by HLTA, specifically those generated by subroutine \sysname{OneIsland} (Algorithm~\ref{alg.oneisland}). It is carried out by the two subroutines \sysname{Pem-Lcm} and \sysname{Pem-Ltm-2l}.

At lines~\ref{alg3.line1} and~\ref{alg3.firstlcm}, \sysname{OneIsland} needs to estimate the parameters of an LCM with three observed variables. It is done using EM. Next, it enters a loop. At the beginning, we have an LCM $m$ for a set $\mathcal{S}$ of variables. The parameters of the LCM have been estimated earlier (line~\ref{alg3.firstlcm} at beginning or line~\ref{alg3.lcm} of previous pass through the loop). At lines~\ref{alg3.pick-x} and~\ref{alg3.pick-w}, \sysname{OneIsland} finds the variable $X$ outside $\mathcal{S}$ that has maximum MI with $\mathcal{S}$, and the variable $W$ inside $\mathcal{S}$ that has maximum MI with $X$.

\begin{algorithm}[t]
\small
\begin{algorithmic}[1]
\State  $Y \gets$ the latent variable of $m$;
\State $\mathcal{S}_1 \gets \{X\} \cup$ two seed variables in $\mathcal{S}$;\label{alg.pem-lcm.line2}
\State While keeping the other parameters fixed, run EM in the part of $m$ that involves $\mathcal{S}_1 \cup Y$ to  estimate $P(X|Y)$.
\State \Return $m$
\end{algorithmic}
\caption{\sysname{Pem-Lcm}($m$, $\mathcal{S}$, $X$, $\mathcal{D}$)}\label{alg.pem-lcm}

\end{algorithm}

\begin{algorithm}[t]
\small
\begin{algorithmic}[1]
\State $Y \gets$ the latent variable of $m$;
\State $m_2 \gets$ model obtained from $m$ by adding $X$ and a new latent variable
$Z$, connecting $Z$ to $Y$, connecting  $X$ to $Z$, and re-connecting   $W$  (connected to $Y$ before) to $Z$;
\State $\mathcal{S}_1 \gets \{W, X\} \cup$ two seed variables\footnotemark in $\mathcal{S}$;
\State While keeping the other parameters fixed, run EM in the part of $m_2$ that involves $\mathcal{S}_1 \cup Y \cup Z$ to only estimate $P(W|Z)$, $P(X|Z)$ and $P(Z|Y)$.
\State \Return $m_2$
\end{algorithmic}
\caption{\sysname{Pem-Ltm-2l}$(m, \mathcal{S} \setminus \{W\}, \{W,X\}, \mathcal{D})$}\label{alg.pem-ltm}
\end{algorithm}

At line~\ref{alg3.lcm}, \sysname{OneIsland} adds $X$ to the $m$ to create a new LCM $m_1$. The parameters of $m_1$ are estimated using the subroutine \sysname{Pem-Lcm} (Algorithm~\ref{alg.pem-lcm}), which is an application of progressive EM. Let us explain \sysname{Pem-Lcm} using the intermediate models shown in Figure~\ref{fig.oneisland}.  Let $m$ be the model shown on the left of Figure~\ref{fig.oneislandc} and $\mathcal{S} = \{\word{nasa},\word{space}, \word{shuttle}, \word{mission}, \word{moon} \}$. The variable $X$ to be added to $m$ is \word{lunar}, and the model $m_1$ after adding \word{lunar} to $m$ is shown on the left of Figure~\ref{fig.oneislandd}.  The only distribution to be estimated is $P(lunar|Y)$, as other distributions have already been estimated. \sysname{Pem-Lcm} estimates the distribution by running EM on a part of the model $m_1$ in Figure~\ref{fig.x} (left), where the variables involved are in rectangles.  The variables \word{nasa} and \word{space} are included in the submodel, instead of other observed variables, because they were the seed variables picked at line~\ref{alg3.s1} of Algorithm~\ref{alg.oneisland}.

\footnotetext{When one of the seed variables is $W$, use the other seed variable and the variable picked at line~\ref{alg3.s2} of Algorithm~\ref{alg.oneisland}.}
{
\begin{figure*}[t]
\begin{center}
 \includegraphics[scale=.2]{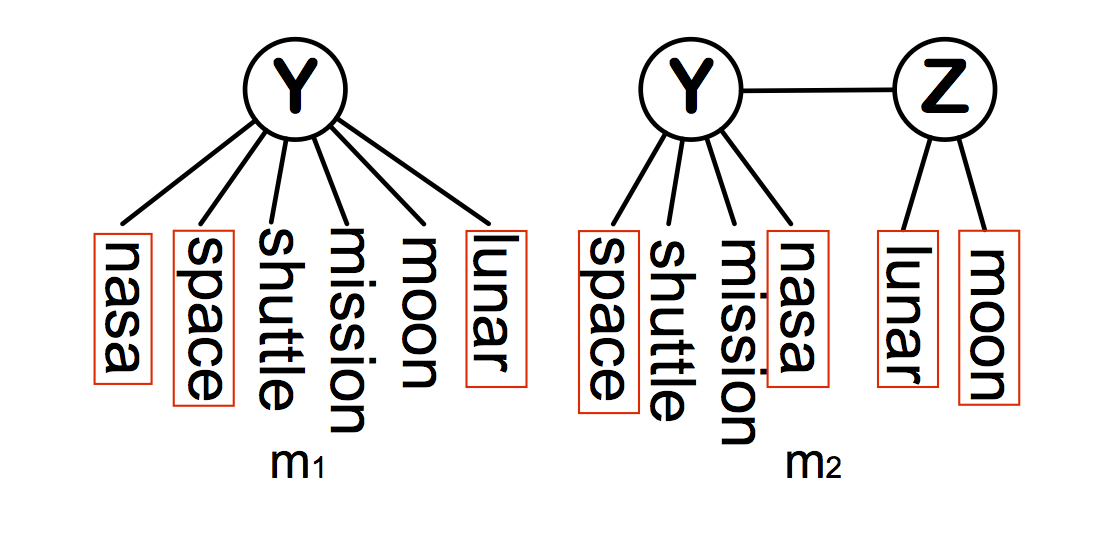}
\end{center}
\caption{ Parameter estimation during island building: To determine whether the variable \word{lunar} should be added to the island $m_1$ in Figure~\ref{fig.oneislandc}, two models are created. We need to estimate only $P(lunar|Y)$ for the model on the left, and $P(Z|Y)$, $P(moon|Z)$ and $P(lunar|Z)$ for the model on the right. The estimation is done by running EM in the parts of the models where the variable names are in rectangles.}
\label{fig.x}
\end{figure*}
}

At line~\ref{alg3.ltm}, \sysname{OneIsland} adds $X$ to the $m$ to create a new LTM $m_2$ with two latent variables. The parameters of $m_2$ are estimated using the subroutine \sysname{Pem-Ltm-2l} (Algorithm~\ref{alg.pem-ltm}), which is also an application of progressive EM. In our running example, let \word{moon} be the variable $W$ that has the highest MI with \word{lunar} among all variables in $\mathcal{S}$. Then the model $m_2$ is as shown on the right hand side of Figure~\ref{fig.oneislandd}. The distributions to be estimated are: $P(Z|Y), P(moon|Z)$ and $P(lunar|Z)$.  \sysname{Pem-Ltm-2l} estimates the distributions by running EM on a part of the model $m_2$ in Figure~\ref{fig.x} (right), where the variables involved are in rectangles.  The variables  \word{nasa} and \word{space} are included in the submodel, instead of \word{shuttle} and \word{mission}, because they were the seed variables picked at line~\ref{alg3.s1} of Algorithm~\ref{alg.oneisland}.

There is also a parameter estimation problem inside the subroutine \sysname{BridgedIslands}. After linking up the islands, the parameters for edges between latent variables must be estimated. We use progressive EM for this task also. Consider the model in Figure~\ref{fig.overall} (a). To estimate $P(Z11|Z12)$, we form a sub-model by picking two children of $Z11$, for instance \word{nasa} and \word{space}, and two children of $Z12$, for instance \word{orbit} and \word{earth}. Then we estimate the distribution $P (Z11|Z12)$ by running EM in the submodel with all other parameters fixed.

\subsection{Complexity Analysis}
\label{sec.complexAnalysis}
Let $n$ be the number of observed variables and $N$ be the sample size. HLTA requires the computation of empirical MI between each pair of observed variables. This takes $O(n^2N)$ time.

When building islands for the observed variables, HLTA generates roughly $2n$ intermediate models.  Progressive EM is used to estimate the parameters of the intermediate models. It is run on submodels with 3 or 4 observed variables. The projection of a dataset onto 3 or 4 binary variable consists of only 8 or 16 distinct cases no matter how large the original sample size is. Hence  progressive EM takes constant time, which we denote by $c_1$, on each submodel. This is the key reason why HLTA can scale up. The data projection takes $O(N)$ time for each submodel. Hence the total time for island building is $O(2n(N+c_1))$.

To bridge the islands, HLTA needs to estimate the MI between every pair of latent variables and runs progressive EM to estimate the parameters for the edges between the islands. A loose upper bound on the running time of this step is $n^2N +n(N+c_1)$.  The total number of variables (observed and latent) in the resulting flat model is upper bounded by $2n$. Inference on the model takes no more $2n$ propagation steps for each data case. Let $c_2$ be the time for each propagation step. Then the hard assignment step takes $O(4nc_2N)$ time. So, the total time for the first pass through the loop in HLTA is $O(2n^2N+3n(N+c_1) + 4nc_2N) = O(2n^2N + 3nc_1+4nc_2N)$, where the term $3nN$ is ignored because it is dominated by the term $4nc_2N$.

As we move up one level, the number of  ``observed'' variables is decreased by at least half. Hence, the total time for the model construction phase is upper bounded by $O(4n^2N + 6nc_1+8nc_2N)$.

The total number of variables (observed and latent) in the final model is upper bounded by $2n$. Hence, one EM iteration takes $O(4nc_2N)$ time and the final parameter optimization steps takes $O(4nc_2N\kappa)$ times.

The total running time of HLTA is $O(4n^2N + 6nc_1+8nc_2N) +O(4nc_2N\kappa)$. The two terms are  the times for model construction phase and the parameter estimation phase respectively.

\section{Dealing with Large Datasets}
\label{sec.big}

We employ two techniques to further accelerate HLTA so that it can handle large datasets with millions of documents. The first technique is downsampling and we use it is to reduce the complexity of the model construction phase. Specifically,  we use a subset of $N'$ randomly sampled data cases instead of the entire dataset and  thereby reduce the complexity to $O(4n^2N' + 6nc_1+8nc_2N')$. When $N$ is very large, we can set $N'$ to be a small fraction of $N$ and hence achieve substantial computational savings. In the meantime, we can still expect to obtain a good structure if $N'$ is not too small. The reason is that  model construction relies on salient regularities of data and those regularities should be preserved in the subset when $N'$ is not too small.

The second technique is stepwise EM \cite{stepEM2000,stepEM2009}. We use it to accelerate the convergence of the parameter estimation process in the second phase, where the task is to improve the values of the parameters $\theta = \{\theta_{ijk}\}$ (Equation~\ref{eq.theta_ijk}) obtained in the model construction phase. While standard EM, a.k.a. {\em batch EM},  updates the parameter once in each iteration,
stepwise EM updates the parameters multiple times in each iteration.

Suppose the data set ${\cal D}$ is randomly divided into equal-sized minibatches ${\cal D}_1$, \ldots, ${\cal D}_B$. Stepwise EM updates the parameters after processing each minibatch. It maintains a collection of auxiliary variables $n_{ijk}$, where are initialized to $0$ in our experiments. Suppose the parameters have been updated $u-1$ times before and the current values are $\theta = \{\theta_{ijk}\}$.  Let  ${\cal D}_b$ be the next minibatch to process. Stepwise EM carries out the updating as follows:
\begin{eqnarray}
\label{eq.sEM-E}
n'_{ijk} &=& \sum_{d \in \mathcal{D}_b} P(V_i=k, pa(V_i)=j|d, m, \theta),\\
\label{eq.sEM-E2}
n_{ijk} &=& (1 - \eta_u) n_{ijk} + \eta_u n'_{ijk},\\
\label{eq.sEM-M}
\theta_{ijk} &=& \frac{ n_{ijk}}{\sum_k n_{ijk}}.
\end{eqnarray}
\noindent Note that equation (\ref{eq.sEM-E}) is similar to (\ref{eq.EM-E}) except that the
  statistics are calculated on the minibatch ${\cal D}_b$ rather than the entire dataset ${\cal D}$. The parameter $\eta_u$ is known as the {\em stepsize} and is given by
$\eta_u = (u + 2 )^{-\alpha}$ and the parameter $\alpha$ is to be chosen the range $0.5 \leq \alpha \leq 1$ \cite{liang2009online}. In all our experiments, we set $\alpha=0.75$.

Stepwise EM is similar to stochastic gradient descent~\cite{bousquet2008tradeoffs} in that it updates the parameters after processing each minibatch. It has been shown to yield estimates of the same or even better quality as batch EM and it converges much faster than the latter \cite{liang2009online}.  As such, we can run it for much fewer iterations than batch EM and thereby substantially reduce the running time. %Alternatively, we can choose to stop stepwise EM after a predetermined number of updates. This way, it can potentially terminate even before it finishes the first iteration, which might be necessary for every large datasets.

\section{Illustration of Results and Practical Issues}
\label{sec.results}

HLTA is a novel method for hierarchical topic detection and, as discussed in the introduction, it is fundamentally different from the LDA-based methods. We will empirically compare HLTA with the LDA-based methods in the next section. In this section, we present the results HLTA obtains on a real-world dataset so that the reader can gain a clear understanding of what it has to offer. We  also discuss two practical issues.

\subsection{Results on the NYT Dataset}
\label{sec.resultsNIPS}
\begin{figure}[!h]
\begin{center}
\centering
\includegraphics[width=16.2cm, angle =90]{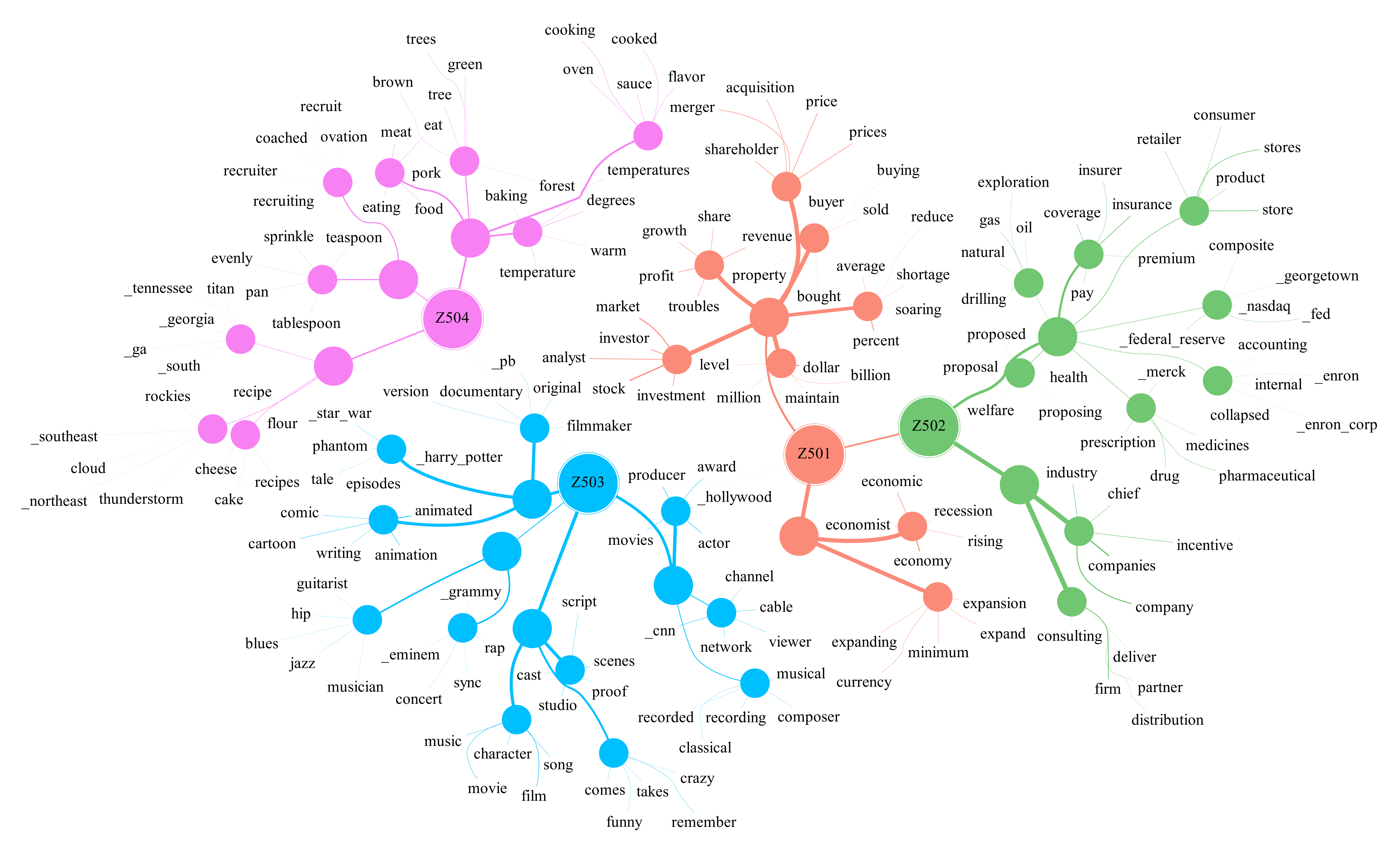}
\caption{\small  Part of the hierarchical latent tree model obtained by HLTA on the NYT dataset:  The nodes $Z501$ to$Z504$ are top-level latent variables. The subtrees rooted at different top-level latent variables are coded using different colors. Edge width represents mutual information.}
\label{fig.nyttree}
\end{center}
\end{figure}

HLTA is implemented in Java. The source code is available online \footnote{ {\tt http://www.cse.ust.hk/{$\sim$}lzhang/topic/index.htm}}, along with the datasets used in this paper and the full details of the results obtained on them. HLTA has been tested on several datasets.
 One of them is the  NYT dataset,
which consists of 300,000 articles published on New York Times between 1987 and
2007\footnote{{\tt http://archive.ics.uci.edu/ml/datasets/Bag+of+Words}}. A vocabulary of 10,000 words was selected using {\em average TF-IDF} \cite{ullman2011mining} for the analysis. The average TF-IDF of a term $t$ in a collection of documents ${\cal D}$ is defined as follows:
\begin{eqnarray}
{\tt tf{-}idf} (t, {\cal D}) &=& \frac{\sum_{d \in {\cal D}} {\tt tf}(t, d) \cdot
{\tt idf}(t, {\cal D})}{|{\cal D}|},
\end{eqnarray}
 \noindent where $| \cdot |$ stands for the cardinality of a set, ${\tt tf}(t, d)$
 is the term frequency of $t$ in document $d$,  and ${\tt idf}(t, {\cal D}) = \log (|{\cal D}|/|\{d \in {\cal D}: t \in d\}|)$ is the inverse document frequency of $t$ in the document collection ${\cal D}$.

 The subset of 10,000 randomly sampled data cases   was used in the model construction phase.
 Stepwise EM was used in the parameter estimation phase and the size of minibatches was set at 1,000. Other parameter settings are given in the next section.
The analysis took around 420 minutes on a desktop machine.

The result is an HLTM with 5 levels of latent variables and 21 latent variables at the top level. Figure~\ref{fig.nyttree} shows a part of the model structure. Four top-level latent variables are included in the figure. The level-4 and level-2 latent variables in the subtrees rooted at the five top-level latent variables are also included. \FloatBarrier  Each level-2 latent variable is connected to four word variables in its subtrees. Those are the word variables that have the highest MI with the latent variable among all word variables in the subtree.

The structure is interesting.  We see that most words in the subtree rooted at $Z501$ are about \emph{economy and stock market}; most words in the subtree $Z502$ are about \emph{companies and various industries}; most words in the subtree rooted at $Z503$ are about \emph{movies and music}; and most words in the subtree rooted at $Z504$ are about \emph{cooking}.

\begin{table}[!t]
{\tt \scriptsize
\caption{Part of topic hierarchy obtained by HLTA on the NYT dataset}
\label{tbl.nyttree}
\begin{tabular}{p{1cm}p{12cm}}
&{\bf 1. [0.20]  economy stock economic market dollar }\\
&\quad 1.1.   [0.20] economy economic economist rising recession\\
&\quad 1.2.   [0.20] currency minimum expand expansion wage\\
&\qquad 1.2.1. [0.20] currency expand expansion expanding euro\\
&\qquad 1.2.2. [0.23] labor union demand industries dependent \\
&\qquad 1.2.3. [0.21] minimum wage employment retirement\\
&\quad ----------------------\\
&\quad 1.3. [0.20]     stock market investor analyst investment      \\
&\quad 1.4. [0.20]     price prices merger shareholder acquisition   \\
&\quad 1.5. [0.20]    dollar billion level maintain million lower\\
&\quad 1.6. [0.20]    profit revenue growth troubles share revenues       \\
&\quad 1.7. [0.20]    percent average shortage soaring reduce\\

&{ \bf 2. [0.22] companies company firm industry incentive}\\
&\quad 2.1.   [0.23] firm consulting distribution partner\\
&\quad 2.2.   [0.23] companies company industry\\
&\quad ----------------------\\
&\quad 2.3. [0.14] insurance coverage insurer pay premium \\
&\quad 2.4. [0.25] store stores consumer product retailer \\
&\quad 2.5. [0.21] proposal proposed health welfare\\
&\quad 2.6. [0.20] drug pharmaceutical prescription \\
&\quad 2.7. [0.07] federal reserve fed nasdaq composite\\
&\quad 2.8. [0.09] enron internal accounting collapsed \\
&\quad 2.9. [0.09]   gas drilling exploration oil natural  \\

\end{tabular}
}
\end{table}

Table~\ref{tbl.nyttree} shows a part of the topic hierarchy extracted from the part of the model shown in Figure~\ref{fig.nyttree}. The topics and the relationships among them are meaningful. For example, the topic 1 is about economy and stock market. It splits into two groups of subtopics, one on economy and another on stock market.  Each subtopic further splits into sub-subtopics. For example, the subtopic 1.2 under economy splits into three subtopics: currency expansion, labor union and minimum wages. The topic 2 is about company-firm-industry. Its subtopics include several types of companies such as insurance, retail stores/consumer products, natural gas/oil, drug, and so on.

\subsection{Two Practical Issues}

Next we discuss two practical issues.

\subsubsection{Broadly vs Narrowly Defined Topics}
\label{sec.BandNtopics}
In HLTA, each latent variable is introduced to model a pattern of probabilistic word co-occurrence. It also gives us a topic, which is a soft cluster of documents. The size of the topic is determined by considering not only the words in the pattern, but all the words in the vocabulary. As such, it conceptually includes two types of documents: (1) documents that contain, in a probabilistic sense, the pattern, and (2) documents that do not contain the pattern but are otherwise similar to those that do. Because of the inclusion of the second type of documents, the topic is said to be \emph{broadly defined}. All the topics reported above are broadly defined.

The size of a widely defined topic might appear unrealistically large at the first glance. For example, one topic detected from the NYT dataset  consists of the words \word{affair}, \word{widely}, \word{scandal}, \word{viewed}, \word{intern}, \word{monica\_lewinsky}, and its size is $0.18$.  Although this seems too large, it is actually reasonable. Obviously, the fraction of documents that contain the seven words in the topic should be much smaller than $18\%$. However, those documents also contain many other words, such as \word{bill} and \word{clinton}, about American politics. Other documents that contain many of those other words are also included in the topic, and hence it is not surprising for the topic to cover $18\%$ of the corpus.  As a matter of fact, there are several other topics about American politics that are of similar sizes. One of them is: \word{corruption campaign political democratic presidency}.

In some applications, it might be desirable to identify {\it narrowly defined topics} --- topics made up of only the documents containing particular patterns. Such topics can be obtained as follows: First, pick a list of words to characterize a topic using the method described in Section  \ref{sec.hltm}; then, form a latent class model using those words as observed variables; and finally, use the model to partition all documents into two clusters. The cluster where the words occur with relatively higher probabilities are designated as the narrow topic. The size of a narrowly defined topic is typically much smaller than that of the widely defined version. For example, the sizes of the narrowly defined version of the two topics from the previous paragraph are 0.008 and 0.169 respectively.

Learning a latent class model for each latent variable from scratch is time-consuming. To accelerate the process, one can calculate the parameters for the latent class model from the global HLTM model, fix the conditional distributions, and update only the marginal distribution of the latent variable a number of times, say 10 times.

\subsubsection{Use of N-grams as Observed Variables}

\begin{table}[!t]
\caption{Topics detected from AAAI/IJCAI papers (2000-15)  that contain the
word "network".}
{\tt \scriptsize
\centering
    \begin{tabular}{p{0.1cm}p{12cm}}
 & \quad  [0.05] neural-network neural hidden-layer layer activation \\
& \quad [0.08]  bayesian-network probabilistic-inference variable-elimination\\
& \quad [0.04] dynamic-bayesian-network dynamic-bayesian slice time-slice \\
& \quad  [0.03] domingos markov-logic-network richardson-domingos\\
& \quad [0.06]  dechter constraint-network freuder consistency-algorithm\\
& \quad [0.09] social-network twitter social-media social tweet\\
&\quad  [0.03] complex-network community-structure community-detection\\
& \quad [0.01] semantic-network conceptnet partof \\
& \quad [0.09] wireless sensor-network remote radar radio beacon\\
& \quad [0.08] traffic transportation road driver drive road-network
\end{tabular}
\label{tbl.network-topics}
}
\end{table}

In HLTMs, each observed variable is connected to only one latent variable.
If individual words are used as observed variables, then each word would appear
in only one branch of the resulting topic hierarchy. This is not reasonable. Take the word ``network" as an example. It can appear in different terms such as ``neural network", ``Bayesian network", ``constraint network", ``social network", ``sensor network", and so on. Clearly, those terms should appear in different branches in a good topic hierarchy.

A method to mitigate the problem is proposed in \cite{poon2016topic}. It first treats individual words as tokens and finds top $n$ tokens with the highest average TF-IDF. Let \sysname{Pick-top-tokens}$({\cal D},n)$ be the subroutine which does that. The method then calculates the TF-IDF values of all 2-grams, and and includes the top $n$ 1-grams and 2-grams with highest TF-IDF values as tokens. After that, the selected 2-grams (e.g., ``social network'') are replaced with single tokens (e.g., ``social-network'') in all the documents and the subroutine \sysname{Pick-top-tokens} $({\cal D},n)$ is run again to pick a new set of $n$ tokens. The process can be repeated if one wants to consider n-grams with $n>2$ as tokens.

The method has been applied in an analysis of papers published at AAAI and IJCAI between 2000 and 2015. Table \ref{tbl.network-topics} shows some of the topics from the resulting topic hierarchy. They all contain the word ``network" and are from different branches of the hierarchy.

\section{Empirical Comparisons}
\label{sec.comparisonWLDA}

We now present empirical results to compare HLTA with LDA-based methods for hierarchical topic detection, including nested Chinese restaurant process (nCRP)~\cite{blei10nested}, nested hierarchical Dirichlet process (nHDP)~\cite{paisley2015nested} and Hierarchical Pachinko allocation model(hPAM)~\cite{mimno2007mixtures}. Also included in the comparisons is CorEx~\cite{versteeg14discovering}. CorEx produces a hierarchy of latent variables, but not a probability model over all the variables. For comparability, we convert the results into a hierarchical latent tree model. \footnote{Let $Y$ be a latent variable and $Z_1$, \ldots, $Z_k$ be its children. Let $Z^{(d)}_i$ be the value of $Z_i$ in a data case $d$. It is obtained via hard assignment if $Z_i$ is a latent variable.
 CorEx gives the distribution $P(Y|Z^{(d)}_1, \ldots, Z^{(d)}_k)$ for data case $d$. Let
 ${\bm 1}^{(d)}(Z_1, \ldots, Z_k)$ be a function that takes value $1$ when $Z_i = Z_i^{(d)}$ for all $i$ and 0 otherwise. Then, the expression $\sum_{d \in {\cal D}}P(Y|Z^{(d)}_1, \ldots, Z^{(d)}_k){\bm 1}^{(d)}(Z_1, \ldots, Z_k)/|{\cal D}|$ defines a joint distribution over $Y$ and $Z_1$ - $Z_k$. From the joint distribution, we obtain  $P(Z_i|Y)$ for each $i$, and also  $P(Y)$ if $Y$ is the root.}

\subsection{Datasets}

Three datasets were used in our experiments. The first one is the NYT dataset mentioned before.
The second one is the 20 Newsgroup (referred to as Newsgroup) dataset\footnote{\tt http://qwone.com/$\sim$jason/20Newsgroups/}.  It consists of 19,940 newsgroup posts.  The third one is the  NIPS dataset, which consists of 1,955 articles published
 at the NIPS conference between 1988 and 1999
 \footnote{{\tt http://www.cs.nyu.edu/{$\sim$}roweis/data.html}}. Symbols, stop words and the words barely occurring were removed for all the datasets.

Three versions of the NIPS dataset and two versions of the Newsgroup dataset were created by choosing vocabularies with different sizes using average TF-IDF. For the NYT dataset, only one version was chosen. So, the experiments were performed on six datasets. Information about them is given in Table~\ref{tbl:datasets}. Each dataset has two versions: a binary version where word frequencies are discarded, and a bags-of-words version where word frequencies are kept. HLTA and CorEx were run only on the binary version because they can only process binary data. The methods nCRP, nHDP and hPAM were run on both versions and the results are denoted as {\em nCRP-bin, nHDP-bin, hPAM-bin, nCRP-bow, nHDP-bow} and {\em hPAM-bow} respectively.

\setlength{\tabcolsep}{4pt}
\begin{table}[!t]
\caption{Information about the datasets used in empirical comparisons.}
  \begin{center}
  \begin{tabular}{l|cccccc}
         &NIPS-1k & NIPS-5k & NIPS-10k& News-1k& News-5k& NYT  \\ \hline
  Vocabulary Size & 1,000 & 5,000 & 10,000& 1000 & 5000 &10,000 \\ \hline
  Sample Size  &1,955 & 1,955 & 1,955 &19,940 &19,940 & 300,000\\ \hline
  \end{tabular}
  \end{center}
\label{tbl:datasets}
\end{table}

\subsection{Settings}
HLTA was run in two modes. In the first mode, denoted as {\em HLTA-batch}, the entire dataset was used in the model construction phased and batch EM was used in the parameter estimation phase. In the second mode, denoted as {\em HLTA-step}, a subset of $N'$ randomly sampled data cases was used in the model construction phase and stepwise EM was used in the parameter estimation phase (see Section \ref{sec.big}). In all experiments,  $N'$ was set at 10,000, the size of minibatch at 1,000, and the parameter $\alpha$ in stepwise EM at 0.75.  HLTA-batch was run on all the datasets, while HLTA-step was run only on NYT and the Newsgroup datasets. HLTA-step was not run on the NIPS datasets because the sample size is too small.
For HLTA-batch, the number $\kappa$  of iterations  for batch EM was set at 50. For HLTA-step,  stepwise EM was terminated after 100 updates.

The other parameters of HLTA (see Algorithm \ref{alg.top}) were set as follows
in both modes: The threshold $\delta$ used in UD-tests was set at $3$;  the upper bound  $\mu$ on island size was set at 15; and the upper bound  $\tau$  on the number of top-level topics was set at $30$ for NYT and 20 for all other datasets. When extracting topics from an HLTM (see Section
\ref{sec.hltm}), we ignored the level-1 latent variables because the topics they give are too fine-grained and often  consist of different forms of the same word (e.g., ``image, images").

 The LDA-based methods nCRP, nHDP and hPAM do not learn model structures. A hierarchy needs to be supplied as input. In our experiment, the height of hierarchy was set at 3, as is usually done in the literature. The number of nodes at each level was set in such way that nCRP, nHDP and hPAM would yield roughly the same total number of topics as HLTA. CorEx is configured similarly. We used program packages and default parameter values provided by the authors for these algorithms.

 All experiments were conducted on the same desktop computer. Each experiment was repeated 3 times so that variances can be estimated.

\subsection{Model Quality and Running Times}

For topic models, the standard way to assess  model quality is to measure  log-likelihood on a held-out test set \cite{blei10nested,paisley2015nested}.
 In our experiments, each dataset was randomly partitioned into a training set with $80\%$ of the data, and a test set with $20\%$ of the data. Models were learned from the training set, and per-document loglikelihood was calculated on the held-out test set. The statistics are shown in Table~\ref{tbl:perll}.
 For comparability, only the results on binary data are included.

 We see that the held-out likelihood values for HLTA are drastically higher than those for
all the alternative methods. This implies that the models obtained by HLTA can predict unseen data much better than the other methods. In addition, the variances are significantly smaller for HLTA than the other methods in most cases.

 \begin{table}[!t]

 \begin{minipage}{1.0\textwidth}
{\rowcolors{2}{white}{gray!20}
%\begin{table}[h]
\caption{Per-document held-out loglikelihood on binary data: Best scores are marked in bold. The sign ``-'' indicates non-termination after 72 hours, and  ``NR'' stands for ``not run'.}
\begin{small}
	\begin{center}
	\begin{tabular}{l|cccccc}
        &	{\scriptsize{NIPS-1k}}&	{\scriptsize{NIPS-5k}}&	{\scriptsize{NIPS-10k}}&	 {\scriptsize{News-1k}}&	 {\scriptsize{News-5k}} &	{\scriptsize{NYT}}\\ \hline
	  HLTA-batch	& {\bf -393$\pm$0} &{\bf -1,121$\pm$1}& {\bf -1,428$\pm$1}& {\bf -114$\pm$0}&{\bf -242$\pm$0} & {\bf -754$\pm$1} \\
     HLTA-step & NR & NR & NR & {\bf -114$\pm$0 } & -243$\pm$0 & { -755$\pm$1}\\
   % nCRP-c   &-2,951$\pm$35 &-5,626$\pm$117& --- &---&--- & --- \\
    nCRP-bin   & -671$\pm$16 &-3,034$\pm$135& --- &---&--- & ---\\
   % nHDP-c  & -4,107$\pm$6 &-7,986$\pm$11 &-9,147$\pm$56 & -329$\pm$1 & -686$\pm$3&-2,059$\pm$8 \\
    nHDP-bin  & -1,188$\pm$1 &-3,272$\pm$3 &-4,001$\pm$11 & -183$\pm$1 & -407$\pm$2&-1,530$\pm$5 \\
    hPAM-bin & -1,183$\pm$ 3 & ---  & --- &-183$\pm$2 & ---& ---\\
    CorEx  &-445$\pm$2 &-1,243$\pm$2 & -1,610$\pm$4 &-149$\pm$1& -323$\pm$4& ---\\
    \end{tabular}
	\end{center}
\end{small}
\label{tbl:perll}
%\end{table}
}

\end{minipage}

\begin{minipage}{1.0\textwidth}

{\rowcolors{2}{white}{gray!20}
\caption{Running times:  The sign ``-'' indicates non-termination after 72 hours, and  ``NR'' stands for ``not run''. }
\begin{small}
  \begin{center}
  \begin{tabular}{l|cccccc}
    Time(min)     & {\scriptsize{NIPS-1k}}& {\scriptsize{NIPS-5k}}&  {\scriptsize{NIPS-10k}}&  {\scriptsize{News-1k}}&  {\scriptsize{News-5k}} &  {\scriptsize{NYT}} \\ \hline
    HLTA-batch & {\bf 5.6$\pm$0.1} &{\bf 86.1$\pm$3.7} &318$\pm$12&  66.1$\pm$2.9 & 432$\pm$39 & 787$\pm$42\\
    HLTA-step & NR & NR & NR &{\bf 9.6$\pm$0.1} &  {\bf 133$\pm$5} & {\bf 421$\pm$17} \\
    nCRP-bin   & 782$\pm$39 &  3,608$\pm$163 &--- &--- &---&--- \\
    nHDP-bin  & 152$\pm$1& 288$\pm$16 & {\bf 299$\pm$9} & 162$\pm$13 & 263$\pm$9 & 430$\pm$42\\
    hPAM-bin &  261$\pm$17& ---  & --- &328$\pm$9& ---& ---\\
    nCRP-bow   & 853$\pm$150 &3,939$\pm$301 &--- &--- &---&--- \\
    nHDP-bow  &379$\pm$14& 416$\pm$49&413$\pm$16 & 250$\pm$81& 332$\pm$16 & 564$\pm$59 \\
    hPAM-bow & 850$\pm$27 & ---  & --- & 604$\pm$19 & ---& ---\\
    CorEx  & 53$\pm$0.2&371$\pm$23&1,190$\pm$9 &779$\pm$34 &4,287$\pm$52& ---\\
    \end{tabular}
  \end{center}
\end{small}
\label{tbl:Time}

}

\end{minipage}
\end{table}

Table~\ref{tbl:Time} shows the running times.
We see that HLTA-step is significantly more efficient than HLTA-batch on large datasets, while there is virtually no decrease in model quality.

Note that all algorithms have parameters that control computational complexity. Thus, running time comparison is only meaningful when it is done with reference to model quality. It is clear from Tables \ref{tbl:perll} and \ref{tbl:Time} that HLTA achieved much better model quality  than the alternative algorithms using comparable or less time.

 %Both HLTA-batch and HLTA-step are much more efficient than nCRP, hPAM and CorEx. As we observed, nCRP and hPAM can take prohibitively long time while dealing with more than hundreds of topics. HLTA-batch was faster than nHDP-bin on the smaller datasets (NIPS-1k, NIPS-5k, News-1k), and slower on the larger datasets (NIPS-10k, News-5k, NYT).  HLTA-step took roughly the same amount of time as nHDP-bin on the largest dataset NYT.
%
%

\subsection{Quality of Topics}

It has been argued that, in general, better model fit does not necessarily imply better topic quality~\cite{chang2009reading}.  It might therefore be more meaningful to compare alternative methods in terms of topic quality directly. We measure topic quality using two metrics.
The first one is  the \emph{topic coherence score} proposed by~\cite{mimno2011optimizing}.
  Suppose a topic $t$ is characterized using a list $W^{(t)}=\{w_1^{(t)}, w_2^{(t)}, \ldots, w_M^{(t)}\}$ of  $M$ words. The coherence score of $t$ is given by:
  \begin{eqnarray}
  {\tt Coherence}(W^{(t)}) = \sum\limits_{i=2}^{M}\sum\limits_{j=1}^{i-1}\log \dfrac{D(w_i^{(t)},w_j^{(t)})+1}{D(w_j^{(t)})},
  \end{eqnarray}
where $D(w_i)$ is the number of documents containing  word $w_i$, and $D(w_i, w_j)$ is the number of documents containing both $w_i$ and $w_j$. It is clear that the score depends on the choices of $M$ and it generally decreases with $M$. In our experiments, we set $M=4$ because some of the topics produced by HLTA have only 4 words and hence the choice of a larger value for $M$ would put other methods at a disadvantage. With $M$ fixed, a higher coherence score indicates a better quality topic.

 \begin{table}[!t]

 \begin{minipage}{1.0\textwidth}

{\rowcolors{2}{white}{gray!20}

\caption{Average topic coherence scores. }
\begin{small}
	\begin{center}
	\begin{tabular}{l|cccccc}
        &	{\scriptsize{NIPS-1k}}&	{\scriptsize{NIPS-5k}}&	{\scriptsize{NIPS-10k}}&	 {\scriptsize{News-1k}}&	 {\scriptsize{News-5k}} &	{\scriptsize{NYT}} \\ \hline
	  HLTA-batch & {\bf-5.95$\pm$0.04}&{\bf -7.74$\pm$0.07}&{\bf -8.15$\pm$0.05}&-12.00$\pm$0.09& -12.67$\pm$0.15 &  -12.09$\pm$0.07 \\
	    %-12.09$\pm$0.07 \\
      HLTA-step & NR & NR & NR &{\bf -11.66$\pm$0.19}& {\bf -12.08$\pm$0.11}& {\bf -11.97$\pm$0.05}\\
    nCRP-bow  & -7.46$\pm$0.31 &-9.03$\pm$0.16 &--- & ---&---& ---  \\
    nHDP-bow  & -7.66$\pm$0.23 & -9.70$\pm$0.19  &-10.89$\pm$0.38 & -13.51$\pm$0.08& -13.93$\pm$0.21& -12.90$\pm$0.16\\
    hPAM-bow &  -6.86$\pm$0.08 & ---  & --- & -11.74$\pm$0.14 & ---& ---\\
    nCRP-bin  & -7.01$\pm$0.37 &-9.83$\pm$0.08 &--- & ---&---& --- \\
    nHDP-bin  & -8.95$\pm$0.11 & -11.59$\pm$0.12  &-12.34$\pm$0.11 & -13.45$\pm$0.05 & -14.17$\pm$0.08& -14.55$\pm$0.07 \\
    hPAM-bin & -6.83$\pm$0.11& ---  & --- &-12.63$\pm$0.06 & ---& ---\\
	CorEx  & -7.20$\pm$0.23& -9.76$\pm$0.48& -11.96$\pm$0.52&-13.49$\pm$1.48 & -14.71$\pm$0.45&--- \\
    \end{tabular}
	\end{center}
\end{small}
\label{tbl:TopicCoherence}

}
\end{minipage}

\begin{minipage}{1.0\textwidth}
{\rowcolors{2}{white}{gray!20}
\caption{Average compactness scores.}
\begin{small}
	\begin{center}
	\begin{tabular}{l|cccccc}
        &	{\scriptsize{NIPS-1k}}&	{\scriptsize{NIPS-5k}}&	{\scriptsize{NIPS-10k}}&	 {\scriptsize{News-1k}}&	 {\scriptsize{News-5k}} &	{\scriptsize{NYT}} \\ \hline
	  HLTA-batch & {\bf0.253$\pm$0.003}&{\bf 0.279$\pm$0.008}&{\bf 0.265$\pm$0.001}& 0.239$\pm$0.010& 0.239$\pm$0.006 & { 0.337$\pm$0.003}\\
	  %0.337$\pm$0.003\\
      HLTA-step & NR & NR & NR &{\bf 0.250$\pm$0.003} & {\bf 0.243$\pm$0.002} & {\bf 0.338$\pm$0.002}\\
    nCRP-bow  & 0.163$\pm$0.003 &0.153$\pm$0.001 &--- & ---&---& ---  \\
       nHDP-bow  & 0.164$\pm$0.005 &0.147$\pm$0.006 &0.138$\pm$0.002 & 0.150$\pm$0.003 & 0.148$\pm$0.004 & 0.250$\pm$0.003\\
    hPAM-bow &0.215$\pm$0.013& --- & --- &0.210$\pm$0.006&---&---\\
    nCRP-bin  & 0.176$\pm$0.005 &0.137$\pm$0.005 &--- & ---&---& --- \\
    nHDP-bin  & 0.119$\pm$0.005 &0.107$\pm$0.003 &0.102$\pm$0.003 & 0.138$\pm$0.003 & 0.134$\pm$0.003 & 0.166$\pm$0.001\\
     hPAM-bin &0.145$\pm$0.008& --- & --- &0.169$\pm$0.010&---&---\\
     CorEx  & 0.243$\pm$0.018 &0.162$\pm$0.013 &0.167$\pm$0.003 & 0.185$\pm$0.012 & 0.156$\pm$0.009 & --- \\
    \end{tabular}
	\end{center}
\end{small}
\label{tbl:Compactness}
}

\end{minipage}
\end{table}

The second metric we use is  the \emph{topic compactness score} proposed by~\cite{chen2016sparse}.
The compactness score of a topic  $t$ is given by:
\begin{eqnarray}
{\tt compactness}(W^{(t)}) = \dfrac{2}{M(M-1)}\sum\limits_{i=2}^{M}\sum\limits_{j=1}^{i-1} S(w_i^{(t)},w_j^{(t)}),
\end{eqnarray}
where $S(w_i, w_j)$ is the similarity between the words $w_i$ and $w_j$ as determined by a \emph{word2vec} model \cite{mikolov2013efficient,DBLP:conf/nips/MikolovSCCD13}. The  \emph{word2vec} model was trained on a part of the Google News dataset\footnote{https://code.google.com/archive/p/word2vec/}. It  contains about 100 billion words and each word is mapped to a high dimensional vector. The similarity between two words is defined as the cosine similarity of the two corresponding vectors. When calculating ${\tt compactness}(W^{(t)})$, words that do not occur in the  \emph{word2vec} model  were simply skipped.

Note that the coherence score is calculated on the corpus being analyzed. In this sense, it is an {\em intrinsic} metric. The intuition is that words in a good topic should tend to co-occur in the documents. On the other hand, the compactness score is calculated on a general and very large corpus not related to the corpus being analyzed. Hence it is an {\em extrinsic} metric. The intuition here is that words in a good topic should be closely related semantically.

Tables~\ref{tbl:TopicCoherence} and \ref{tbl:Compactness} show the average topic coherence and topic compactness scores of the topics produced by various methods. For LDA-based methods, we reported their scores on both datasets of binary and bags-of-words versions. There is no distinct advantage of choosing either version. We see that the scores for the topics produced by HLTA are significantly higher than the those obtained by other methods in all cases.

\subsection{Quality of Topic Hierarchies}

 \begin{table}[!h]
\caption{A part of topic hierarchy by nHDP.}
{\tt \scriptsize
\begin{tabular}{p{2cm}p{12cm}}

&{\bf 1. company business million companies money  }\\
&  \quad 1.1. economy economic percent government  \\ %mixture\\
&  \quad 1.2. percent   stock   market  analyst quarter  \\
&  \qquad   1.2.1. stock fund market  investor investment \\
&  \qquad   1.2.2.  economy rate rates fed economist  \\
&  \qquad   1.2.3. company quarter million sales analyst\\
&  \qquad   1.2.4.  travel ticket airline flight traveler\\
&  \qquad   1.2.5.  car ford sales vehicles\_chrysler\\
&  \quad 1.3. computer technology system software   \\
&  \quad 1.4. company   deal million billion  stock  \\
&  \quad 1.5. worker job union employees contract  \\
&  \quad 1.6. project million plan official area\\
\end{tabular}
}
\label{tbl.nytnhdp}
\end{table}

There is no metric for measuring the quality of topic hierarchies to the best of our knowledge, and it is difficult to come up with one.  Hence, we resort to manual comparisons.

The entire topic hierarchies produced by HLTA and nHDP on the NIPS and NYT datasets can be found at the URL mentioned at the beginning of the previous section. Table~\ref{tbl.nytnhdp} shows the part of the hierarchy by nHDP that corresponds to the part of the hierarchy by HLTA shown in Table~\ref{tbl.nyttree}. In the HLTA hierarchy, the topics are nicely divided into three groups, {\it economy}, {\it stock market}, and {\it companies}.  In Table~\ref{tbl.nytnhdp}, there is no such clear division. The topics are all mixed up. The hierarchy does not match the semantic relationships among the topics.

Overall, the topics and topic hierarchy obtained by HLTA are more meaningful than those by nHDP.

%

%Table~\ref{tbl:TopicCoherence}
%\begin{table}[!htb]
%\caption{Hierarchy coherence scores}
%\begin{small}
%	\begin{center}
%	\begin{tabular}{l|ccccc}
%          &	{\scriptsize{Nips-1k}}&	{\scriptsize{Nips-5k}}&	 {\scriptsize{Nips-10k}}&	 {\scriptsize{News-1k}}&	 {\scriptsize{News-5k}} \\ \hline
%	{\scriptsize{PEM-HLTA}}	& -19.76&-31.09 &-37.50&-37.06&-37.06 \\ \hline
%    HLTA   &-21.51 &-31.50  &--- & -37.55&---\\ \hline
%    hLDA   & & & & & \\ \hline	
%    nHDP  & & &-30.35 & & \\ \hline
%    CorEx  & & & & & \\ \hline
%    \end{tabular}
%	\end{center}
%\end{small}
%\label{tbl:TopicCoherence}
%\end{table}
%HLTA hierarchy differ from LDA hierarchy: Nature (NIPS), meaningfulness  (NIPS ), and coherence (NIPs and other)

\section{Conclusions and Future Directions}
\label{sec.conclusion}
We propose a novel method called HLTA for hierarchical topic detection. The idea is to model patterns of word co-occurrence and co-occurrence of those patterns using a hierarchical latent tree model. Each latent variable in an HLTM represents a soft partition of documents and the document clusters in the partitions are interpreted as topics. Each topic is characterized using the words that occur with high probability in documents belonging to the topic and occur with low probability in documents not belonging to the topic. Progressive EM is used to accelerate parameter learning for intermediate models created during model construction, and stepwise EM is used to accelerate parameter learning for the final model. Empirical results show that HLTA outperforms nHDP, the state-of-the-art  method for hierarchical topic detection based on LDA, in terms of overall model fit and quality of topics/topic hierarchies, while takes no more time than the latter.

HLTA treats words as binary variables and each word is allowed to appear in only one branch of a hierarchy. For future work, it would be interesting to extend HLTA so that it can handle count data and that a word is allowed to appear in multiple branches of the hierarchy. Another direction is to further scale up HLTA via distributed computing and by other means.

\section*{Acknowledgments}
\label{sec.ack}
We thank John Paisley for sharing the nHDP implements with us, and we thank Jun Zhu for valuable discussions. Research on this article was supported by Hong Kong Research Grants Council under grants 16202515 and 16212516,  and the Hong Kong Institute of Education under project RG90/2014-2015R.

\section*{References}

\bibliography{topic-bib}
\bibliographystyle{elsarticle-num}
\end{document}